\documentclass{article}

\PassOptionsToPackage{numbers, compress}{natbib}

\usepackage[main, final]{neurips_2026}
\usepackage{graphicx}
\usepackage{multirow}
\usepackage{subcaption}
\usepackage{colortbl}
\usepackage{xcolor}
\usepackage{wrapfig}
\usepackage{amsmath}

\usepackage[utf8]{inputenc} 
\usepackage[T1]{fontenc}    
\usepackage{hyperref}       
\usepackage{url}            
\usepackage{booktabs}       
\usepackage{amsfonts}       
\usepackage{nicefrac}       
\usepackage{microtype}      
\usepackage{xcolor}         
\usepackage{tcolorbox}
\tcbuselibrary{skins, breakable, listings}

\newtcolorbox{promptbox}[2][]{%
    enhanced,
    breakable,
    colback=white,       
    colframe=gray!30!black, 
    colbacktitle=gray!40!black, 
    coltitle=white,      
    fonttitle=\bfseries\sffamily\large,
    title={#2},          
    sharp corners,       
    boxrule=1.0pt,       
    left=6pt, right=6pt, top=6pt, bottom=6pt,
    #1
}

\newcommand{\placeholder}[1]{\textcolor{blue!70!black}{\texttt{<#1>}}}

\title{InterSketch: An Interleaved Reasoning Model with Self-correcting Visual Sketch and Stepwise Reward}

\newcommand{\ours}{InterSketch}

\author{%
    \textbf{Zhiwei Ning$^{1,*}$} \quad
    \textbf{Wenwen Tong$^{2,*,\Delta}$} \quad
    \textbf{Xiangli Kong$^{2,*}$} \quad
    \textbf{Shengnan Ma$^{2,*}$} \quad
    \textbf{Ziyi Shang$^{2}$} \\
    \textbf{Jingcheng Ni$^{2}$} \quad
    \textbf{Tao Hu$^{2}$} \quad
    \textbf{Yong Xien Chng$^{2}$} \quad
    \textbf{Jixuan Ying$^{2}$} \quad
    \textbf{Zehuan Wu$^{2}$} \\
    \textbf{Hanming Deng$^{2}$} \quad
    \textbf{Jie Yang$^{1}$} \quad
    \textbf{Yuanjie Zheng$^{3}$} \quad
    \textbf{Wei Liu$^{1,\dagger}$} \quad
    \textbf{Lewei Lu$^{2,\dagger}$} \\[6pt]
    $^1$Shanghai Jiao Tong University \quad
    $^2$SenseTime Research \quad
    $^3$Shandong Normal University \\[3pt]
    {\small \texttt{zwning@sjtu.edu.cn} \quad \texttt{tongwenwen1@sensetime.com}} \\[3pt]
    {\small \texttt{weiliucv@sjtu.edu.cn} \quad \texttt{luotto@sensetime.com}} \\
    {\small $^*$Equal contribution \quad $^\Delta$Project leader \quad $^\dagger$Corresponding author}
  }

\begin{document}

\maketitle


  {
    \centering
    \includegraphics[width=1.0\columnwidth]{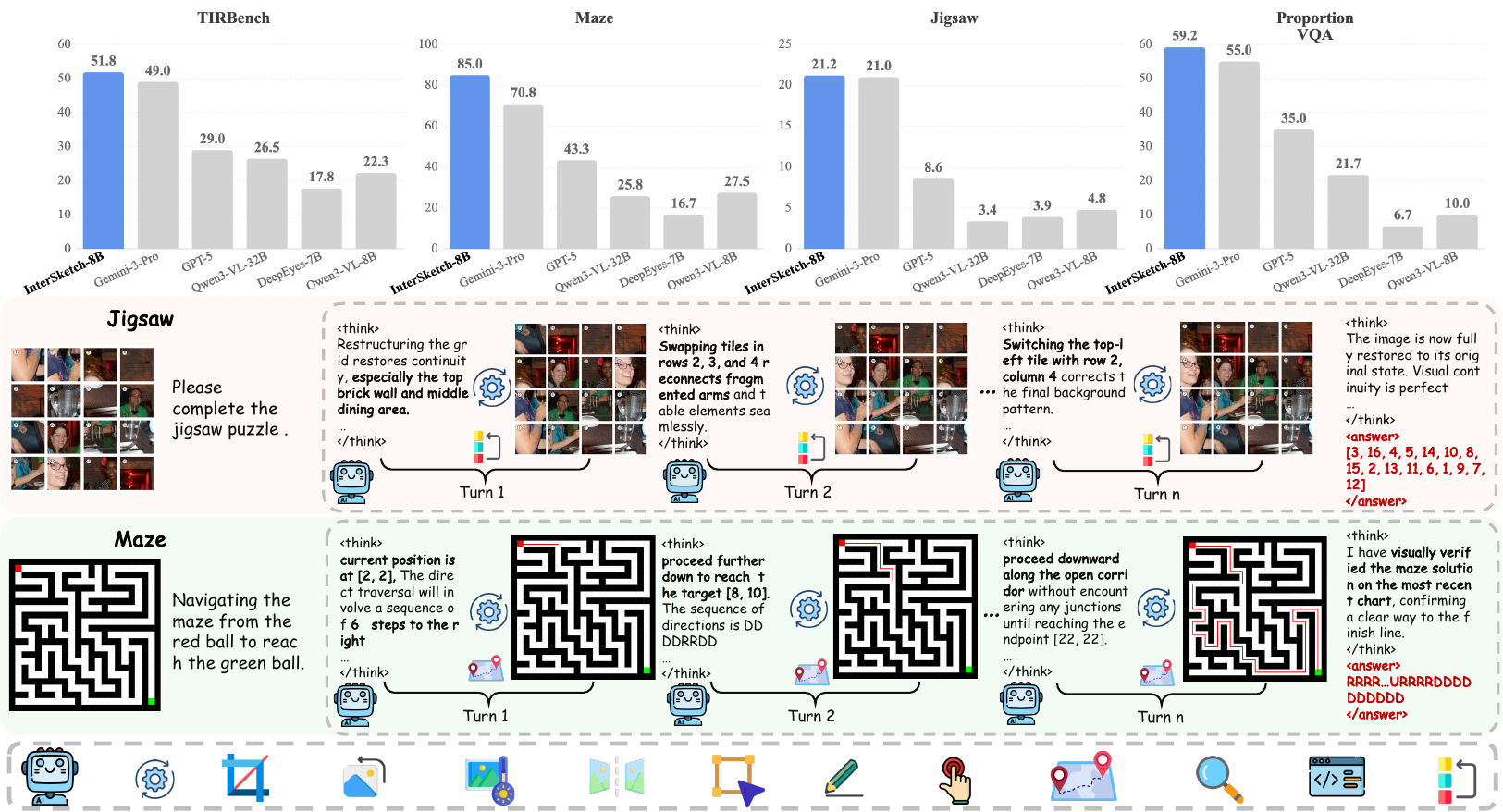} 
    \captionof{figure}{Performance on the interleaved reasoning tasks. Our \ours{} consistently achieves state-of-the-art performance and even surpasses proprietary models such as Gemini-3-Pro. Our pipeline iteratively generates textual rationales and invokes visual tools to generate updated sketches, which serve as explicit scaffolds for complex logical deduction.}
    \label{fig:teaser}
  }
  \vskip 0.1in

\begin{abstract}

While vision-language models (VLMs) have exhibited multi-turn visual reasoning capabilities, their reasoning trajectories remain relatively shallow and are dominated by a text-centric paradigm, limiting their applicability to complex visual challenges. In contrast, human-like thought typically involves long-horizon reasoning with an interleaved visual-textual chain-of-thought (VT-CoT). To bridge this gap, we introduce \ours{}, an interleaved reasoning model to enhance the VT-CoT capability via self-correcting and stepwise reward mechanisms. \ours{} dynamically generates intermediate visual sketches using external tools and interleaves them with textual reasoning, enabling effective perception and logical reasoning over long-horizon visual understanding tasks. Specifically, in the first cold-start stage, we propose a synthesized high-quality interleaved VT-CoT dataset and include a reflection mechanism to enable the model's capability in multi-turn interleaved reasoning and self-correction. In the subsequent reinforcement learning (RL) stage, we design a stepwise reward mechanism to mitigate the sparsity of reward signals inherent in end-only supervision over long-horizon reasoning. Extensive experiments on visual reasoning benchmarks demonstrate the effectiveness of \ours{}, even outperforming proprietary models such as Gemini-3-Pro.


\end{abstract}
\vspace{-1.0em}
\section{Introduction}
\vspace{-0.5em}


The reasoning and perception abilities of vision-language models (VLMs)~\cite{Qwen3-VL, wang2025internvl3} have advanced significantly across a wide range of visual understanding tasks, largely driven by techniques such as chain-of-thought (CoT)~\cite{wei2022chain} and reinforcement learning (RL)~\cite{guo2025deepseek}.
Meanwhile, \textit{thinking-with-images}~\cite{su2025thinking, OpenAI2025Thinking, gu2025thinkmorph} approaches further enhance these capabilities by enabling models to invoke external tools such as image crop for detailed visual analysis~\cite{OpenAI2025Thinking, lai2025mini}. 
Despite these advances, most existing multi-modal reasoning approaches still rely primarily on text-centric CoT and relatively shallow reasoning trajectories with simple external tool usage~\cite{zheng2025deepeyes, zhang2025thyme}. 
As a result, they struggle to produce human-like solutions to complex visual understanding problems that require mental simulation~\cite{shi2025realunify}, especially those demanding multi-step, long-horizon, and interleaved visual–textual chain-of-thought (VT-CoT)~\cite{li2025imagine}, such as the spatial reasoning and state-evolution tasks illustrated in Fig.~\ref{fig:teaser}.
Therefore, it is critical to build an agentic VLM with strong reasoning ability and human-like interleaved VT-CoT to effectively address complex and diverse visual understanding tasks in real-world scenarios.

The potential of \textit{thinking-with-images} paradigm with interleaved image and text reasoning has been underscored by models such as OpenAI-o3 \cite{OpenAI2025Thinking} to improve the visual understanding capabilities.  
MVoT \cite{li2025imagine} proposes the multimodal visualization-of-thought to handle the complex spatial reasoning tasks through generating image visualizations within the reasoning traces. ThinkMorph \cite{gu2025thinkmorph} explores the multimodal interleaved chain of thought reasoning by the iterative coordination between the language and vision for a general visual task.
However, these unified generative-understanding methods exhibit two notable limitations: their reasoning trajectories are relatively shallow, typically involving only a single intermediate image, and their performance is heavily dependent on the unified model’s image-generation quality.
In contrast, tool-generated intermediate sketches integrate more seamlessly with the reasoning process in VLMs by decoupling visual perception from image generation, thereby avoiding dependence on generative quality. DeepEyes \cite{zheng2025deepeyes} and Mini-o3 \cite{lai2025mini} incorporate image manipulation to construct intermediate visual observations and improve reasoning via RL, but their reliance on simple image cropping limits the depth of their reasoning trajectories. DeepEyeV2 \cite{hong2025deepeyesv2}, CodeVision \cite{guo2025thinking}, and SenseNova-MARS~\cite{chng2025sensenova} extend this direction with more flexible multi-tool frameworks and RL algorithms toward agentic VLMs. However, their image manipulation capabilities remain limited to basic operations, and their reward functions rely solely on end-only supervision without providing feedback on intermediate steps. These gaps motivate agentic VLMs that support long-horizon interleaved VT-CoT together with flexible and compositional multi-tool invocation, enabling robust reasoning for challenging visual understanding tasks.

To this end, we introduce \ours{}, an agentic VLM designed to elevate visual reasoning from passive perception to active, interleaved VT-CoT. Unlike traditional approaches that rely on static inputs or end-to-end image generation, \ours{} dynamically interacts with a deterministic library of tools---such as cropping, rotation, and geometric annotation---to produce intermediate visual sketches. These sketches serve as explicit cognitive scaffolds, allowing the model to externalize its mental simulation and ground subsequent reasoning in verified visual evidence. In this way, \ours{} enables tackling general and challenging visual tasks that require long-horizon reasoning.

Specifically, to bridge the gap in long-horizon visual reasoning capabilities of existing models, we construct a large-scale corpus of multi-step synthetic data. To further enable self-reflection during inference, we deliberately embed error-recovery episodes into the supervised fine-tuning (SFT) data synthesis, including both parameter errors and reasoning path errors, which teach the model to diagnose failures and take corrective actions. Building upon these curated dataset, we adopt a two-stage training pipeline to equip the model with tool-augmented multi-step visual reasoning capabilities. Notably, to mitigate the sparsity of end-only rewards inherent in previous RL paradigms, we introduce a stepwise reward mechanism that calculates progressive improvement based on the tool-call types to enhance interleaving efficiency. Overall, our method achieves state-of-the-art performance on challenging visual reasoning benchmarks and surpasses proprietary models like Gemini-3-Pro, as illustrated in the top panel of Figure~\ref{fig:teaser}, while simultaneously preserving strong foundational capabilities on general understanding tasks. In summary, our contributions are threefold:

\begin{itemize}

\item We propose \ours{}, a VT-CoT schema VLM that employs visual scaffolds for long-horizon interleaved reasoning, post-trained on high-quality synthetic dataset with explicit reflection mechanisms that endow the model with self-correction capabilities.
\item We introduce a hybrid stepwise reward mechanism for RL training that assigns category-specific reward functions based on the type of tool invocation, enabling more efficient credit assignment and policy optimization in tool-augmented reasoning.
\item Our model achieves state-of-the-art performance on challenging visual reasoning benchmarks, surpassing proprietary models such as Gemini-3-Pro, while also generalizing to general understanding tasks with consistent performance gains.

\end{itemize}



\section{Related Works}
\label{sec:related_works}

\subsection{Interleaved Visual-Textual Reasoning}
Conventional VLMs typically encode visual inputs in a single pass and then conduct text-only reasoning, which limits their ability to handle problems that require iterative inspection and refinement of visual evidence, especially for complex spatial and temporal reasoning~\cite{yue2024mmmu, lu2023mathvista, wu2024vsp}. 
Recent advances in interleaved multimodal reasoning aim to overcome this limitation by enabling models to process sequences where image and text tokens are interwoven, allowing information to be exchanged across modalities throughout the reasoning process~\cite{alayrac2022flamingo, li2025otter}.
In addition, a growing line of research introduces explicit intermediate visual representations as part of the reasoning trace. MVoT \cite{li2025imagine} proposes multimodal visualization-of-thought, generating visualizations within reasoning trajectories to better support challenging spatial reasoning tasks. ThinkMorph~\cite{gu2025thinkmorph} studies iterative coordination between language and vision for general visual tasks via an interleaved chain-of-thought, but the reasoning trajectories are often shallow. 
Overall, existing interleaved reasoning methods demonstrate the promise of visual-textual co-evolution, yet how to produce controllable and reusable intermediate visual artifacts that support long-horizon reasoning remains an open question.

\subsection{Tool-Augmented Agentic Visual Reasoning}
Tool-augmented frameworks instantiate agentic visual reasoning by allowing models to actively manipulate visual inputs or construct intermediate observations with external tools, rather than relying solely on passive perception~\cite{wu2023visual, yang2023mm, zhang2025thyme}.
The thinking-with-images paradigm, highlighted by OpenAI-o3~\cite{OpenAI2025Thinking}, further emphasizes interleaved image and text reasoning with iterative visual operations, resembling human mental simulation through active perception and deliberation~\cite{su2025thinking, su2025openthinkimg}. 
Early attempts, such as Pixel Reasoner~\cite{wang2025pixel} and DeepEyes~\cite{zheng2025deepeyes} provide the open-source demonstration that RL can incentivize such active visual behaviors and improve fine-grained performance.
However, training agentic VLMs for complex multi-turn tool use remains challenging. 
Pure RL often struggles with a vast action space and sparse reward signal, making credit assignment across multi-step tool invocations difficult~\cite{wang2025pixel, zhang2025thyme, lai2025mini, zhou2025reinforced}.
Mini-o3~\cite{lai2025mini} reports that pure RL fails to elicit deep multi-step search trajectories and adopts multi-stage training.
Recent methods such as DeepEyesV2~\cite{hong2025deepeyesv2}, CodeVision~\cite{guo2025thinking}, and SenseNova-MARS~\cite{chng2025sensenova} extend multi-tool frameworks with RL, but their manipulations remain limited to basic operations, and their end-only reward signals lack intermediate-step supervision, hindering generalization and effective credit assignment in long-horizon reasoning.

\section{Method}

\begin{figure*}[t]
  \centering
  \includegraphics[width=\textwidth]{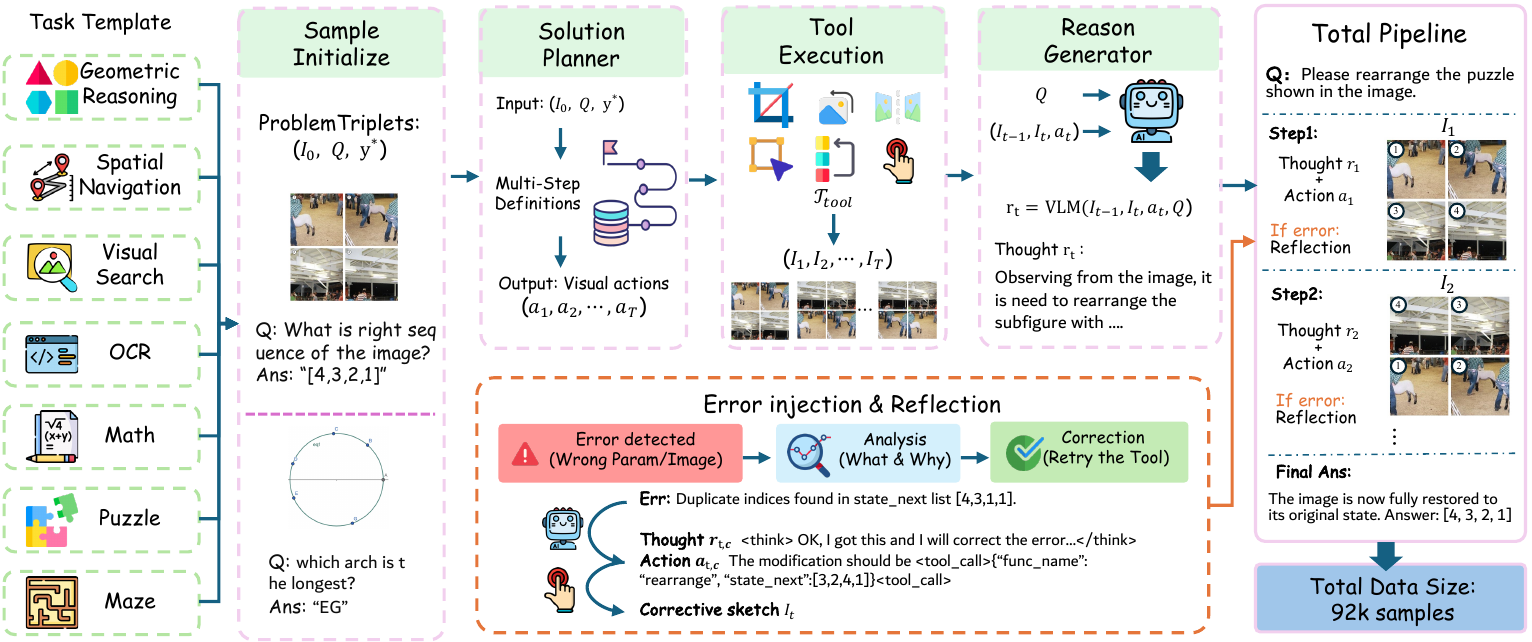}
  \caption{Cold-start data synthesis pipeline. The pipeline automatically generates high-quality interleaved reasoning trajectories through combining sample initialization, solution planner, tool execution, reason generation, and error injection with reflection mechanisms.}
  \label{fig:data_pipeline}
\end{figure*}

\subsection{Problem Definition}
\label{subsec:problem_definition}

We formulate the visual understanding task not as a static mapping from an image to an answer, but as a dynamic, agentic process of interleaved VT-CoT. 
Formally, given a data sample $x = (I_0, Q)$ consisting of the initial visual image $I_0$ and question $Q$, the goal is to generate a final answer $y$, which is similar to the ground-truth answer $y^\ast$. 
Unlike previous paradigms that reason directly on the initial image input, we posit that solving complex visual challenges requires generating a sequence of intermediate visual sketches $\mathcal{V}_t = \{I_1, \dots, I_T\}$, where $T$ indicates the total number of reasoning steps and $I_t$ denotes the intermediate sketch generated at step $t$. 
Specifically, \ours{} employs the external tool to generate the intermediate visual sketches.
Given the historical context $h_t$ at reasoning step $t$, the policy $\pi_\phi(r_t, a_t | h_t)$ predicts a reasoning trajectory consisting of the thought $r_t$ and tool-use action $a_t \in \mathcal{A}_{\text{tool}}$.
Following the generation of the action $a_t$, the environment invokes the execution module $\mathcal{T}_{\text{tool}}$ to generate a new visual sketch:
\begin{equation}
    I_t = \mathcal{T}_{\text{tool}}(I_{t-1}, a_t).
\end{equation}
The details of the tool-set and execution module $\mathcal{T}_{\text{tool}}$ are provided in the supplementary.
The thought $r_t$ and action $a_t$, combined with the new sketch $I_t$, are explicitly appended to the history $h_{t+1} \leftarrow h_t \cup (r_t, a_t, I_t)$. This cycle forms the interleaved VT-CoT step, and the whole trajectory $\tau$ is defined as:
\begin{equation}
\label{eq:traj}
    \tau = [(I_0, q), \underbrace{(r_1, a_1, I_1)}_{\text{Step 1}}, \underbrace{(r_2, a_2, I_2)}_{\text{Step 2}}, \dots, y],
\end{equation}
where $\tau[:t]=h_t$ denotes the history trajectory generated before step $t$, and $y$ is the predicted answer.

\subsection{Data Synthesis Pipeline with Reflection and Self-correction Mechanism}
\label{subsec:data_pipeline}

A critical bottleneck in long-horizon tool-augmented VLMs is the absence of large-scale, process-annotated interleaved reasoning data. 
In this section, we introduce an automated data synthesis pipeline to generate diverse and high-quality interleaved reasoning trajectories.

\begin{figure*}[t]
  \centering
  \includegraphics[width=\textwidth]{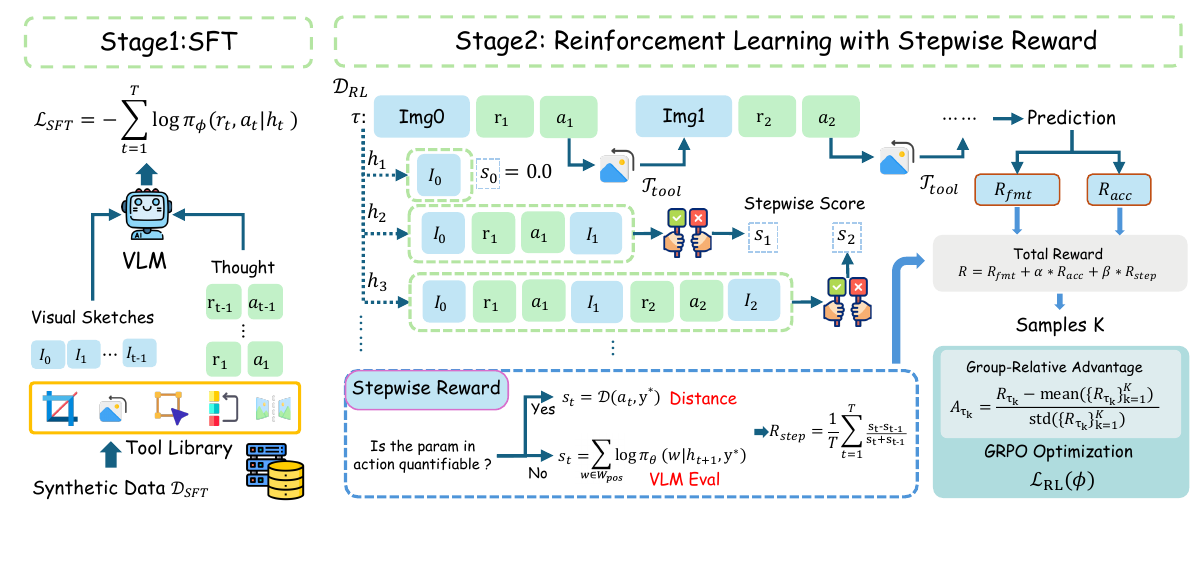}
  \vspace{-2.0em}
  \caption{Two-stage training framework of \ours{}. SFT employs cold-start supervised fine-tuning on our synthetic interleaved reasoning dataset. The policy is enhanced within the RL stage via the stepwise reward, which evaluates the necessity and validity of each tool invocation.}
  \label{fig:training_framework}
\end{figure*}

\paragraph{Cold-start Data with Reflection and Self-correction.}
The cold-start data synthesis pipeline consists of four stages, as illustrated in Figure~\ref{fig:data_pipeline}. 
We first initialize the problem triplet $(I_0, Q, y^*)$ from procedurally generated task templates. 
Second, a solution planner decomposes the problem into multiple steps and outputs a sequence of visual actions $(a_1, a_2, \ldots, a_T)$, where each action $a_t$ indicates a tool invocation along with its parameters. 
For example, visual search tasks yield progressive cropping actions, whereas geometric reasoning tasks produce auxiliary line-drawing actions.
Then, we apply a pre-defined tool function $\mathcal{T}_{tool}$ to manipulate the current visual sketch $I_{t-1}$ based on action $a_t$ and generate the next sketch $I_t$, yielding a sequence of intermediate sketches $(I_1, I_2, \cdots, I_T)$. Subsequently, we synthesize the underlying thought process by feeding  $I_{t-1}$, $I_t$, $a_t$, and $Q$ into a VLM and prompting it to explain why the action $a_t$ is appropriate to transform the current sketch $I_{t-1}$ into $I_t$ to address the question $Q$, which is defined as:
\begin{equation}
  r_t = \text{VLM}(I_{t-1},I_t,a_t,Q).
\end{equation}
Finally, we treat $(r_t, a_t, I_t)$ as an interleaved reasoning step, thereby constructing VT-CoT trajectories as in Eq.~\ref{eq:traj}.
These trajectories form the foundation of our synthetic dataset for cold-start SFT.

Real-world reasoning is rarely linear, and errors are unavoidable. Therefore, to equip VLMs with error-recovery capability, we inject error episodes into training by intentionally corrupting tool-call parameters in a subset of synthesized trajectories. When the executor detects invalid parameters, such as the crop coordinates exceeding image bounds, it returns an error signal without updating the visual state. The VLM is then prompted to output the failure and revise the tool call accordingly.
For the reflection mechanism injected in the step $t$, the process is defined as:
\begin{equation}
\small
  \tau_{\text{ref},t} = \tau[{:}t] \cup [(r_{t,e}, a_{t,e}; \textit{Err}), (r_{t,c}, a_{t,c}; I_t)] \cup \tau[t{+}1{:}],
\end{equation}
where $r_{t,e}$ and $a_{t,e}$ are the wrong thought and action, respectively, and $\textit{Err}$ defines error warning or sketch returned by tool function. $r_{t,c}$ and $a_{t,c}$ are the correct thought and action, while $I_t$ is the right sketch. 
Leveraging such data, the VLM learns to monitor tool-execution status and to adaptively correct parameter specifications upon detecting violations.
Crucially, during the training, we mask the loss on the erroneous response segment $(r_{t,e}, a_{t,e})$ to prevent the model from imitating incorrect outputs. 
In total, we construct 92K samples to form the cold-start dataset $\mathcal{D}_{SFT}$.

\paragraph{RL Training Data for Capability Improvement.}
In contrast to the cold-start SFT data with a complete interleaved reasoning trajectory, the RL training data is composed of only triplet  $(I_0, Q, y^*)$ instances. 
We utilize the same sample initialization as the cold-start phase to generate a candidate pool, and implement a filtering strategy to further enhance the data quality. 
Specifically, for each instance, we perform eight rollouts using the model after the SFT stage and preserve only those samples where the policy model achieves a success rate between 1/8 and 7/8 to construct the final RL training dataset $\mathcal{D}_{RL}$ with 36K samples.

\subsection{VT-CoT Schema Training and Enhancement}
\label{subsec:training_framework}

We adopt a progressive two-stage training pipeline to equip the model with interleaved visual–textual reasoning capabilities, as shown in Figure~\ref{fig:training_framework}. The first cold-start SFT stage establishes foundational tool-use protocols and reasoning patterns using the synthesized cold-start dataset. 
In the subsequent RL stage, we enhance the model's policy by introducing a stepwise reward mechanism to enhance efficiency and accuracy, thereby boosting overall performance on interleaved reasoning.

\paragraph{SFT for Foundational VT-CoT Schema.}
The cold-start phase addresses the fundamental challenge that pre-trained VLMs, despite their strong perceptual and linguistic capabilities, lack the latent knowledge of when and how to invoke visual tools. Therefore, we guide the VLM to imitate the reasoning process and tool invocation patterns on the synthetic corpus $\mathcal{D}_{\text{SFT}}$ via SFT. Our training strategy differentiates between regular and reflection trajectories. For regular trajectories, we apply standard next-token prediction to maximize the log-likelihood across all tokens:
\begin{equation}
  \mathcal{L}_{\text{SFT}}(\phi) = \mathbb{E}_{\tau \sim \mathcal{D}_{\text{SFT}}} \Bigl[ -\sum_{t=1}^{T} \log \pi_\phi(r_t, a_t \,|\, h_t) \Bigr].
\end{equation}
For reflection trajectories containing error recovery episodes, we mask the erroneous reasoning steps to prevent the model from learning incorrect thinking patterns. Specifically, when a trajectory includes an error at step $t$ followed by reflection and correction, we exclude the erroneous tokens $(r_{t,e}, a_{t,e})$ from loss computation by setting the loss weight for those steps to zero. This masking strategy ensures the model learns from the reflection and correction process $(r_{t,c}, a_{t,c})$ without internalizing incorrect intermediate reasoning. The supervised phase establishes the ability to generate structured tool-call actions and interpret their reasoning, providing a strong foundation for subsequent RL.

\paragraph{RL within Stepwise Reward.}

We further optimize tool-use capabilities for complex scenarios during the RL stage. Prior approaches, such as group relative policy optimization (GRPO)~\cite{shao2024deepseekmath}, typically rely on sparse rewards at the end of a trajectory. To overcome this limitation, we introduce a multimodal interleaved stepwise reward mechanism that densities supervision by evaluating the effectiveness of intermediate sketches after each tool invocation.

The core challenge in stepwise supervision is quantifying the performance of an intermediate tool invocation. 
We categorize tasks into two distinct types based on the properties of their action trajectories. 
The first type comprises tasks where the distance between the parameters of the intermediate action $a_t$ and the ground-truth answer is explicitly computable, such as maze navigation. 
The second type involves scenarios where the relationship between the parameter trajectory and the final answer cannot be explicitly formulated.
In the former setting, we define the stepwise score $s_t$ using a metric function $\mathcal{D}$, such as the cosine similarity or the minimum edit distance, to measure the alignment between the generated action trajectory and the ground-truth answer $y^*$.
In contrast, for tasks lacking explicitly quantifiable trajectories, we employ an evaluator VLM $\pi_\theta$ to assess the effectiveness of the tool invocation.
Specifically, we condition $\pi_\theta$ on the historical context $h_{t+1}$ after the current step and the ground-truth answer $y^*$. Then, we prompt the model to determine whether the current visual sketch $I_t$ suffices to deduce the correct answer, followed by summing the log-probabilities of the tokens belonging to the affirmative label set $\mathcal{W}_{\text{pos}}$. Overall, the stepwise score $s_t$ can be formulated as:
\vspace{-0.5em}
\begin{equation}
s_t = 
\left\{
\begin{array}{c l}
    \mathcal{D}(a_t, y^*), & \text{if } a_t \in \mathcal{A}_\text{quant} \\[2ex] 
    \displaystyle \sum\limits_{w \in \mathcal{W}_\text{pos}} \log \pi_{\theta} (w | h_{t+1}, y^*), & \text{otherwise}
\end{array}
\right.
\label{eq:qe_6}
\end{equation}
where $\mathcal{A}_\text{quant}$ is the quantifiable action set, and $s_0$ is initialized to 0. Furthermore, we calculate the final stepwise reward for a trajectory $\tau$ as the average relative improvement between consecutive scores, which is defined as:
\begin{equation}
R_{\text{step}}(\tau) = \frac{1}{T} \sum_{t=1}^{T} \frac{s_t - s_{t-1}}{s_t + s_{t-1}}.
\end{equation}
To guide the policy towards both correct final answers and high-quality intermediate reasoning, we aggregate the sparse format reward, accuracy reward, and the dense stepwise reward into a composite function for the entire trajectory $\tau$, and define the total reward as:
\begin{equation}
  R(\tau) = R_{\text{fmt}}(\tau) + \alpha \cdot  R_{\text{acc}}(\tau) + \beta R_{\text{step}}(\tau), 
\end{equation}
where $R_{\text{fmt}}(\tau)$ is the format reward ensuring the strict compliance with the tool-usage protocol, $R_{\text{acc}}(\tau)$ denotes the final-answer correctness score. We set $\alpha = 0.7$ and $ \beta =0.3 $ to achieve the trade-off between the task completion and process rigor. 
We optimize the policy model $\phi$ using the GRPO strategy. 
For each input query, we sample a group of $K$ trajectories $\{\tau_1, \dots, \tau_K\}$ and compute the advantage for each trajectory $\tau_{k}$ by normalizing its reward against the group mean:
\begin{equation}
\small
\bar{A}_{k} =
\frac{
R(\tau_k) - \operatorname{mean}\!\left(\left\{ R_{\tau_{k}} \right\}_{k=1}^{K}\right)
}{
\operatorname{std}\!\left(\left\{ R_{\tau_{k}} \right\}_{k=1}^{K}\right)
}.
\end{equation}
With the normalized advantages, the final optimization policy is given by:
\begin{equation}
    \mathcal{L}_{\text{RL}}(\phi) = \mathbb{E} \Bigl[ \min\bigl( \rho_k(\phi) \bar{A}_{k},\; \text{clip}(\rho_k(\phi), 1{-}\epsilon_{low}, 1{+}\epsilon_{up}) \bar{A}_{k} \bigr) \Bigr].
\end{equation}

where $\rho_k(\phi) = \frac{\pi_\phi(\tau_k)}{\pi_{\text{old}}(\tau_k)}$ is the probability ratio, $\epsilon_{low}$ and $\epsilon_{up}$ define the lower and upper clipping ratios. 
By integrating the stepwise evaluations into the advantage estimation, our model enables to distinguish effective and redundant tool invocations within a trajectory, leading to more precise credit assignment and faster convergence than end-only reward.

\section{Experiments}
\label{sec:exps}

\subsection{Implementation Details}
\paragraph{Experimental Setups.}
\ours{} is trained on top of Qwen3-VL-8B \cite{Qwen3-VL} using a two-stage pipeline: cold-start SFT followed by RL, implemented with SWIFT \cite{zhao2025swift} and veRL \cite{sheng2025hybridflow} frameworks, respectively.
In the cold-start stage, we freeze the vision encoder and fine-tune the language model for two epochs with a batch size of 256 and a learning rate of 2e-5.
To support long-context multimodal inputs and interleaved visual–textual reasoning trajectories, we set the maximum sequence length to 32,768 tokens.
In the RL stage, we also freeze the vision encoder and employ the GRPO algorithm with a group size of 8 and a global batch size of 128. Furthermore, we employ GPT-4o as our default MLLM evaluator, defined in Eq. \ref{eq:qe_6}.
The optimization uses clipping parameters
$\epsilon_{up}=0.28$ and $\epsilon_{low}=0.2$. 
We utilize a constant learning rate of $1 \times 10^{-5}$ without applying Kullback-Leibler (KL) divergence regularization or entropy regularization. 
To maximize the training efficiency, we implement asynchronous rollouts, capping the multi-turn interaction at 15 turns and the response length at 32,768 tokens. 

\paragraph{Evaluation Benchmarks.}
To comprehensively evaluate our model's performance across diverse visual reasoning paradigms, we employ four distinct benchmarks. We primarily utilize TIR-Bench \cite{li2025tir} to assess agentic thinking-with-images capabilities. Comprising 13 tasks ranging from visual search to geometric reasoning, it directly tests the model's ability to autonomously invoke tools and manipulate visual states over multi-turn interactions. 
For fine-grained spatial logic, we examine performance on the visual spatial planning (VSP) benchmark \cite{wu2024vsp}, which challenges the model to execute sequential decision-making in complex spatial navigation such as mazes.
Additionally, we evaluate GEU subtasks drawn from multimodal unified benchmarks, including RealUnify \cite{shi2025realunify} and Uni-MMMU \cite{zou2025uni}, with a focus on scenarios where image manipulation acts as a functional step to improve understanding. Furthermore, we also evaluate on several understanding benchmarks (such as BLINK \cite{fu2024blink}, MMStar\cite{chen2024we}, and MMMU\cite{yue2023mmmu}) to demonstrate the generalization of \ours{}.
\paragraph{Baselines.} We compare \ours{} against several strong baselines, including proprietary models such as GPT-4o \cite{hurst2024gpt}, GPT-5 \cite{openai2025gpt5}, Gemini-2.5-Flash \cite{comanici2025gemini}, and Gemini-3-Pro \cite{gemini3pro}, as well as open-source models such as  InternVL3.5 \cite{wang2025internvl3}, Qwen3-VL \cite{Qwen3-VL}, and DeepEyes \cite{zheng2025deepeyes}.

\subsection{Main Results}
\label{subsec:main_results}

We present the primary evaluation results on the TIR-Bench \cite{li2025tir} and other challenging visual reasoning benchmarks. Table~\ref{tab:tirbench} summarizes the performance comparison on TIR-Bench across diverse tasks.
\ours{}-8B achieves a SOTA average accuracy of 51.8\%, significantly outperforming both open-source and proprietary models. 
Compared to the basic model Qwen3-VL-8B, \ours{}-8B demonstrates a substantial improvement of nearly 30\%, validating the efficacy of our tool-augmented interleaved CoT framework in unlocking the visual reasoning potential of VLM. Notably, \ours{} surpasses leading proprietary models such as Gemini-3-Pro (49.0\%) and o3-TU (46.0\%) in average performance. We observe particularly strong results in tasks requiring precise spatial manipulation and long-horizon planning. For instance, in the maze navigation task, \ours{}-8B achieves 85.0\% accuracy, outperforming the second-best model Gemini-3-Pro by 14.2\%. 
Similarly, in the rotation game and OCR tasks, \ours{}-8B achieves 97.3\% and 75.0\%, respectively. These results highlight the effectiveness of our long-horizon interleaved VT-CoT paradigm in handling geometric transformations and spatial reasoning challenges that are difficult for current VLMs.

To provide a more comprehensive assessment, we extend our evaluation to general VLM benchmarks beyond tool-augmented reasoning tasks, as depicted in Table~\ref{tab:und_bench}. The empirical results demonstrate that our method consistently outperforms the base model Qwen3-VL-8B by a substantial margin across all benchmarks, with improvements ranging from +2.0\% to +10.8\%. Notably, \ours{} also surpasses recent interleaved image-text approaches such as ThinkMorph \cite{gu2025thinkmorph} and DeepEyeV2\cite{hong2025deepeyesv2}, confirming the effectiveness of our model. This substantiates that the reasoning capabilities acquired through our reflection and stepwise reward mechanisms are broadly transferable and not strictly restricted to scenarios requiring explicit tool invocation.

\begin{table*}[t]
\caption{Comparison on the TIR-Bench. The accuracy (\%) across individual sub-tasks and the overall average is reported. \textbf{Bold} indicates the best results, while \underline{underlined} denotes the second best.}
\label{tab:tirbench}
\centering
\resizebox{\textwidth}{!}{
\begin{tabular}{lc|ccccccccccc}
\toprule
\multirow{2}{*}{Model} & \multirow{2}{*}{Avg} & \multirow{2}{*}{Maze} & \multirow{2}{*}{Color} & Word & Jigsaw & \multirow{2}{*}{OCR} & Rotation & Proportion & Spot & Visual & Symbolic & Low-Light   \\
& & & & Search & Game & & Game & VQA & Diff. & Search & Reasoning & VQA  \\
\midrule
\multicolumn{3}{l}{\textit{Open-source Methods}} & & & & & & & & & & \\
Qwen3-VL-8B & 22.3 & 27.5 & 31.0 & 4.0 & 4.8 & 48.3 & 20.0 & 10.0 & 13.7 & 55.8 & 16.0 & 28.0  \\
Qwen3-VL-32B & 26.5 & 25.8 & 35.0 & 11.0 & 3.4 & 53.0 & 17.3 & 21.7 & 23.5 & 59.2 & 12.0 &  32.0  \\
Qwen3-VL-235B & 24.3 & 13.3 & 34.0 & 14.0 & 6.8 & \underline{65.0} & 14.7 & 17.5 & 19.9 & 54.2 & 16.0 & 38.0  \\
InternVL3-8B & 16.9 & 33.3	& 23.0 & 2.0 & 4.5 & 0.0 & 17.3	& 11.7 & 16.6 & 36.7 & 6.0 & 22.0	 \\
InternVL3.5-8B & 17.3 & 17.5 & 25.0	& 7.0 & 4.8 & 1.7 & 9.3 & 22.5 & 17.3 & 32.5 & 22 & 26.0 \\
InternVL3.5-38B & 18.3 & 11.7 & 31.0 & 11.0 & 3.7 & 3.3 & 17.3 & 17.5 & 11.0 & 35.0	& 26.0 & 34.0  \\
DeepEyes-7B & 17.8 & 16.7 & 22.0 & 1.0 & 3.9 & 41.7 & 12.0 & 6.7 & 19.9 & 50.8 & 19.9 & 16.0 \\
Thyme-7B & 20.8 & 22.5 & 29.0 & 0.0 & 4.6 & 50.0& 13.3 & 6.5 & 12.9 & 54.2 & 42.9 & 16.0  \\
\midrule
\multicolumn{3}{l}{\textit{Proprietary Models}} & & & & & & & & & &   \\
GPT-4o & 17.3 & 20.0 & 26.0 & 0.0 & 6.2 & 10.0 & 20.0 & 22.5 & 19.4 & 35.0 & 10.0 & 26.0 \\
GPT-4.1 & 19.8 & 25.0 & 36.0 & 2.0 & 5.0 & 16.6 & 20.0 & 15.0 & 20.6 & 36.7 & 24.0 & 24.0  \\
GPT-5 & 30.7 & 50.8 & 45.0 & 5.0 & 8.8 & 11.7 & 34.7 & 32.5 & \underline{34.9} & 42.5 & 28.0 & 40.0 \\
o3-TU & 46.0 & 42.5 & \textbf{55.0} & \underline{64.0} & 16.4 & 53.3 & \underline{77.3} & 31.7 & \textbf{41.0} & 57.5 & \textbf{66.0} & \underline{42.0}  \\
Gemini-2.5-Pro & 29.0 & 24.2 & 44.0 & 12.0 & 10.4 & 25.0 & 30.7 & 21.7 & 28.5 & 58.3 & 34.0 &  \underline{42.0} \\
Gemini-3-Pro & \underline{49.0} & \underline{70.8} & 49.0 & 43.0 & \underline{21.0} & 33.3 & 58.3 & \underline{55.0} & 29.9 & \textbf{65.8} & \underline{64.0} & \textbf{66.0}  \\

\midrule
\rowcolor{green!10} \ours{}-8B & \textbf{51.8} & \textbf{85.0} & \underline{52.0} & \textbf{71.0} & \textbf{21.2} & \textbf{75.0} & \textbf{97.3} & \textbf{59.2} & 28.4 & \underline{63.3} & 42.0	& 32.0 \\
\bottomrule
\end{tabular}
}
\end{table*}

\begin{table*}[t]
\caption{Comparison on general understanding benchmarks. The accuracy on each task is reported, including visual search (V* and HR-Bench) and knowledge reasoning (BLINK, MMMU, etc.).}
\label{tab:und_bench}
\centering
\resizebox{0.9\textwidth}{!}{
\begin{tabular}{l|ccccccc}
\toprule
Method & V* & HR-Bench 4K & HR-Bench 8K & BLINK & RealWorldQA & MMStar & MMMU \\
\midrule
Qwen2.5-VL-7B & 76.4 & 68.8 & 65.3 & 55.9 & 68.2 & 51.7 & 57.9 \\
ThinkMorph-7B & 67.0 & 54.0 & 46.7 & 60.1 & 44.5 & 56.9 & 52.7 \\
DeepEyeV2-7B & 81.8 & 77.9 & 73.8 & -- & -- & -- & -- \\ 
Thyme-7B & 82.2 & 77.0 & 72.0  & 56.4 & 70.2 & -- & -- \\
Qwen3-VL-8B  & 73.0 & 70.0 & 68.5 & 61.0 & 61.7 & 55.7 & 63.0 \\ 
\midrule
\rowcolor{green!10} InterSketch (Ours) & \textbf{83.8} & \textbf{78.5} & \textbf{77.8} & \textbf{63.0} & \textbf{70.5} & \textbf{57.7} & \textbf{68.6} \\
\rowcolor{gray!15} \textit{+$\Delta$(Qwen3-VL-8B)} & \textit{+10.8} & \textit{+8.5} & \textit{+9.3} & \textit{+2.0} & \textit{+8.8} & \textit{+2.0} & \textit{+5.6} \\
\bottomrule
\end{tabular}
}
\vspace{-1.0em}
\end{table*}

Table~\ref{tab:unified_bench} summarizes the strong performance of \ours{} on other challenging benchmarks that demand advanced spatial reasoning and generation-enhanced understanding, including VSP \cite{wu2024vsp}, 
\begin{wraptable}{t}{0.5\textwidth}
\centering
\caption{Performance comparison on visual reasoning benchmarks. Our InterSketch is compared against both open-source and proprietary models in terms of accuracy.}
\label{tab:unified_bench}
\resizebox{\linewidth}{!}{
\begin{tabular}{l|ccc}
\toprule
\multirow{2}{*}{Model} & \multirow{2}{*}{VSP} & UniMMMU & \multirow{2}{*}{RealUnify} \\
& & Maze &\\
\midrule
\multicolumn{2}{l}{\textit{Open-source Models}} & \\
InternVL3.5-8B  & 25.2 & 2.7 & 27.0 \\ 
InternVL3.5-38B & 50.8 & 2.7 & 25.8 \\ 
Qwen3-VL-8B & 32.8 & 0.6 & 22.8 \\
Qwen3-VL-30B-A3B & 38.5 & 0.7 & 31.0 \\ 
\midrule
\multicolumn{2}{l}{\textit{Proprietary Models}} & \\
GPT-4o & 44.4 & 0.7 & 27.5 \\ 
Gemini-2.5-Flash & 59.3 & 10.7 & 25.3 \\
\midrule
\rowcolor{green!10} \ours{}-8B & \textbf{73.2} & \textbf{12.0} & \textbf{37.8} \\
\bottomrule
\end{tabular}
}
\vspace{-1.5em}
\end{wraptable}
UniMMMU-Maze \cite{zou2025uni}, and RealUnify \cite{shi2025realunify}. On the VSP benchmark, \ours{} achieves 73.2\% accuracy, significantly outperforming the proprietary model Gemini-2.5-Flash (59.3\%) by a margin of 13.9\%, demonstrating strong capabilities in visual spatial planning. On UniMMMU-Maze and RealUnify benchmarks, \ours{} reaches 12.0\% and 37.8\%, substantially surpassing the Qwen3-VL-30B-A3B model with 11.3\% and 6.8\%, respectively. These consistent improvements across diverse benchmarks highlight the broad effectiveness of the VT-CoT schema in bridging the gap between passive visual perception and active multimodal reasoning.

\subsection{Ablation Study}
\vspace{-0.5em}

\paragraph{Impact on the Reflection Strategy.}
Table~\ref{tab:ref_abla} validates the efficacy of incorporating reflection trajectories into the SFT dataset. The model trained with reflection data consistently outperforms the baseline, achieving notable gains in Spot Difference (18.9\% to 29.9\%) and Proportion VQA (42.5\% to 48.3\%). These results confirm that exposing the agent to explicit error-recovery scenarios fosters intrinsic self-correction capability, which effectively mitigates cascading failures during long-horizon reasoning and enhances overall robustness.
\vspace{-1.0em}

\paragraph{Performance on Different Dataset Scales.}
Table~\ref{tab:scale_abla} underscores the critical role of data volume in the SFT stage. Utilizing a minimal subset of 1K samples in $\mathcal{D}_{\text{SFT}}$ yields a modest average accuracy of 26.3\%, indicating insufficient exposure to tool usage protocols. Conversely, scaling the dataset to 10K triggers a substantial performance surge to 43.7\%. 
Notably, the improvement moderates beyond this point; increasing the data from 50K to the full set yields a marginal gain of only 1.0\%. 
This plateau suggests the model has largely saturated its core tool-use skills, so further expanding the synthetic corpus yields only marginal gains.

\begin{table*}[t]
\begin{minipage}[t]{0.48\linewidth}
\renewcommand{\arraystretch}{0.9}
\caption{ Ablation on the reflection mechanism during the SFT stage. Accuracy (\%) of several sub-tasks in TIR-Bench are reported. }
\label{tab:ref_abla}
\centering
\resizebox{\textwidth}{!}{
\begin{tabular}{c|cccc}
\toprule
SFT Data & Jigsaw & Spot Diff. & Math & Pro. VQA \\
\midrule
w/o Ref & 13.3 & 18.9 & 20.0 & 42.5 \\
\rowcolor{green!10} w/ Ref & \textbf{14.5} & \textbf{29.9} & \textbf{22.3} & \textbf{48.3} \\
\bottomrule
\end{tabular}
}
\end{minipage}
\hfill
\begin{minipage}[t]{0.50\linewidth}
\renewcommand{\arraystretch}{0.9}
\caption{Ablation on the scales of the SFT dataset.}
\label{tab:scale_abla}
\centering
\resizebox{\textwidth}{!}{
\begin{tabular}{l|cccc}
\toprule
\multirow{2}{*}{Data Scale} & TIR-Bench & Word & \multirow{2}{*}{Jigsaw} & Rotation \\
& Avg. & Search & & Game \\
\midrule
1K & 26.3 & 47.0 & 4.5 & 20.0 \\
10K & 43.7 & 68.0 & 7.1 & 80.0 \\
50K & 45.0 & 71.0 & 12.9 & 94.7 \\
\rowcolor{green!10} Overall & \textbf{46.0} & \textbf{72.0} & \textbf{14.5} & \textbf{97.7} \\
\bottomrule
\end{tabular}
}
\end{minipage}
\end{table*}

\begin{table*}[t]
\begin{minipage}[t]{0.48\linewidth}
\renewcommand{\arraystretch}{0.9}
\caption{Ablation on different training stages and reward strategies. ``MT.'' and ``CN.'' represent the metal tracking and cognitive navigation.}
\label{tab:stepwise_abla}
\centering
\resizebox{\textwidth}{!}{
\begin{tabular}{l|cc|cccc}
\toprule
\multirow{2}{*}{Stage} & \multicolumn{2}{c|}{Reward} & \multicolumn{2}{c}{TIR-Bench} & \multicolumn{2}{c}{RealUnify} \\
& base & stepwise & Maze & Jigsaw & MT. & CN. \\
\midrule
Baseline & & & 27.5 & 4.8 & 8 & 17 \\
SFT & & & 50.6 & 14.5 & 19 & 29 \\
RL & \checkmark & & 80.8 & 19.4 & 24 & 29 \\
\rowcolor{green!10} RL & \checkmark & \checkmark & \textbf{85.0} & \textbf{21.2} & \textbf{26} & \textbf{32} \\
\bottomrule
\end{tabular}
}
\end{minipage}
\hfill
\begin{minipage}[t]{0.50\linewidth}
\renewcommand{\arraystretch}{0.9}
\caption{Effectiveness of interleaved visual-tool chain-of-thought. The performance is compared on TIR-Bench across different training stages.}
\label{tab:nec_tool}
\centering
\resizebox{\textwidth}{!}{
\begin{tabular}{l|c|ccc}
\toprule
Method & Avg & Maze & Jigsaw & Pro. VQA \\
\midrule
Text-Only-SFT & 34.3 & 23.1 & 9.2 & 43.1 \\
\rowcolor{green!10} InterSketch-SFT & \textbf{46.0} & \textbf{50.6} & \textbf{14.5} & \textbf{48.0} \\
Text-Only-RL & 38.1 & 26.7 & 12.2 & 48.8 \\
\rowcolor{green!10} InterSketch-RL & \textbf{51.8} & \textbf{85.0} & \textbf{21.2} & \textbf{59.2} \\
\bottomrule
\end{tabular}
}
\end{minipage}

\end{table*}

\begin{wrapfigure}{r}{0.6\textwidth}
\centering
\includegraphics[width=\linewidth]{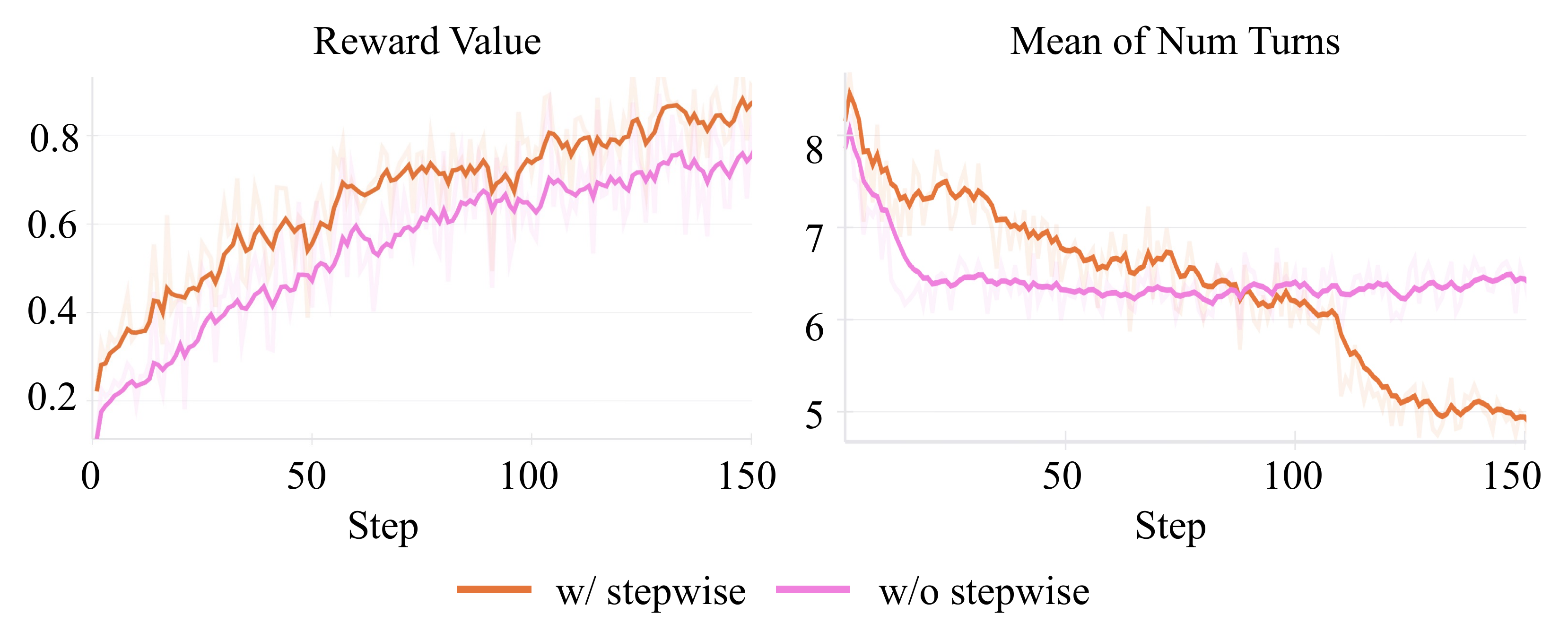}
\caption{Comparison between the w/ and w/o stepwise strategy on the reward score and num turns.}
\label{fig:training_abla}
\end{wrapfigure}

\paragraph{Effectiveness of Stepwise Reward.}
Table~\ref{tab:stepwise_abla} and Figure~\ref{fig:training_abla} provide empirical validation for the training and reward strategies. During SFT, our synthetic data endows the model with foundational visual scaffolding reasoning capabilities, yielding substantial improvements over the baseline model (Qwen3-VL-8B). In the RL stage, while basic end-based reward significantly outperforms the SFT baseline (e.g., boosting Maze accuracy from 50.6\% to 80.8\%), the integration of stepwise reward yields substantial additional gains. 
Specifically, accuracy on the Maze task climbs to 85.0\%, and Jigsaw performance reaches 21.2\%, consistently surpassing the outcome-only variant. Beyond accuracy, the training dynamics in Figure~\ref{fig:training_abla} reveal a critical efficiency advantage: the stepwise supervision not only reaches higher values in the reward but also markedly reduces the average number of interaction turns as the step increases. This indicates that the model learns to prune redundant steps, confirming that explicitly evaluating intermediate visual states mitigates credit assignment difficulties and steers the policy toward reasoning trajectories that are both accurate and computationally efficient.


\paragraph{Impact of Visual-Tool Interleaved Reasoning.}
To demonstrate that a tool-based approach is both justified and necessary for complex visual tasks, we compare it against a text-only reasoning schema, as shown in Table \ref{tab:nec_tool}. Specifically, we strip away the text fields related to sketch descriptions and tool invocations, retaining only the pure logical reasoning context. As the results indicate, relying solely on text-only data leads to a substantial performance degradation in both SFT and RL settings (a drop of 11.7\% and 13.7\%, respectively). This underscores the critical necessity of our sketch-based, multi-tool-calling paradigm in complex visual reasoning scenarios like TIR-Bench.

\vspace{-1.0em}
\section{Conclusion}
\label{sec:conlusions}
\vspace{-1.0em}

We introduce \ours{}, a framework that empowers vision-language models with interleaved visual-textual chain-of-thought reasoning through tool-generated intermediate sketches. By combining a scalable cold-start data synthesis pipeline equipped with explicit reflection mechanisms and a novel stepwise reward for reinforcement learning that densifies supervision at each tool invocation, \ours{} achieves state-of-the-art performance on TIR-Bench and multiple challenging visual reasoning benchmarks, surpassing leading proprietary models such as Gemini-3-Pro. Nevertheless, our approach also has several limitations: the multi-turn rollout process during RL training incurs substantial computational overhead, and all experiments have been conducted at the 8B model scale, leaving scaling behavior unexplored. Future work will investigate dynamic tool discovery, more efficient RL training strategies, and extensions to larger model architectures.

\bibliographystyle{unsrtnat}
\bibliography{example_paper}
\clearpage
\appendix

\section{More Comparisons}

 As summarized in Table \ref{tab:comparison_previous}, we compare the average number of tool calls per sample. Prior works employ single-step or sparse tool calls restricted to narrow domains: OpenThinkIMG \cite{su2025openthinkimg} averages fewer than 0.2 tool calls per sample on chart-only tasks; VTool-R1 \cite{wu2025vtool} invokes at most one tool per instance; and DeepEyes-v2 \cite{hong2025deepeyesv2} and DeepSketcher \cite{zhang2025deepsketcher} performs 1–3 independent calls without chaining intermediate outputs. In contrast, InterSketch executes 4–6 chained invocations per sample, where each intermediate sketch is explicitly fed back as the visual context for subsequent reasoning steps. This closed-loop interleaving constitutes a qualitatively different paradigm—one that enables long-horizon reasoning through iterative visual scaffolding rather than isolated tool augmentation.

\begin{table*}[t]
\caption{Comparison with previous tool-augmented visual reasoning methods.}

\label{tab:comparison_previous}
\centering
\resizebox{\columnwidth}{!}{
\begin{tabular}{l|ccccc}
\toprule
Method & Avg. Tool Calls & Task Diversity & Multi-Step Chain & Stepwise Reward & Reflection \\
\midrule
OpenThinkIMG & $<$0.2 & Chart-only & No & No & No \\
VTool-R1 & $\leq$1 & Chart-only & No & No & No \\
DeepEyes-v2 & 1--2 & General VQA & Limited & No & No \\
DeepSketcher & 1--3 & Drawing & Limited & No & No \\
\rowcolor{green!10} InterSketch (Ours) & 3--6 & 12 task types & Yes & Yes & Yes \\
\bottomrule
\end{tabular}
}
\end{table*}

\section{Tool Definitions}

We formally define the specific tool functions $\mathcal{T}_{\text{tool}}$ utilized within the Actor module, categorized by their operational domains.

\noindent \textbf{$<$Image.Crop$>$ for Fine-grained Analysis.} 
We utilize the \texttt{crop\_image} function to enable the model to focus on specific regions of interest. The function accepts a bounding box $B = [x_1, y_1, x_2, y_2]$ with coordinates normalized to the range $[0, 1000]$, alongside an image index. It extracts the pixel data defined by $B$ from the source image $v_t$, returning a zoomed-in sub-region $v_{t+1}$ to facilitate detailed feature extraction.

\noindent \textbf{$<$Image.Rotate$>$ for Geometric Adjustment.} 
To handle orientation-sensitive tasks, we define the \texttt{rotate\_image} function. It takes the current visual state $v_t$ and a rotation angle $\theta$ (in degrees) as input. Positive values of $\theta$ induce a clockwise rotation. The function outputs the transformed image $v_{t+1}$, allowing the agent to correct the viewpoint or align objects.

\noindent \textbf{$<$Image.Brighten$>$ for Photometric Adjustment.} 
To improve visibility in varying lighting conditions, we utilize the \texttt{brighten\_image} function. It adjusts the pixel intensity of $v_t$ by a scalar factor $\alpha$. Values of $\alpha > 1$ increase brightness, while $0 < \alpha < 1$ decrease it. The function returns the photometrically adjusted image $v_{t+1}$.

\noindent \textbf{$<$Visual.BBox$>$ for Object Localization.} 
We define the \texttt{draw\_bbox} function to visualize specific regions or detected objects. It overlays a bounding box defined by normalized coordinates $B = [x_1, y_1, x_2, y_2]$ onto the current image $v_t$. This visual markup helps the model verify its internal localization predictions explicitely.

\noindent \textbf{$<$Visual.Line$>$ for Trajectory Visualization.} 
We adapt the \texttt{draw\_line} function to sketch linear connections. It takes a coordinate quadruple $[x_1, y_1, x_2, y_2]$ representing the start and end points. An optional extension parameter allows the line to be projected outward as a dashed line, facilitating geometry-related reasoning.

\noindent \textbf{$<$Grid.Route$>$ for Maze Navigation.} 
Designed for pathfinding tasks, the \texttt{route\_drawer} function visualizes a path on an $N \times N$ grid overlay. The input consists of a sequence of grid coordinates $G = \{p_1, \dots, p_k\}$ where each $p_i = (r, c)$ denotes the row and column indices. The function connects these points sequentially with a red line on the image $v_t$ to represent the agent's planned route.

\noindent \textbf{$<$Image.Rearrange$>$ for Jigsaw.} 
To manipulate shuffled patches in the Jigsaw task, we define the \texttt{rearrange\_tiles} function to verify spatial arrangements. Specifically, the function takes the grid dimensions $(R, C)$, the current tile permutation $S_{curr}$, and the target permutation $S_{next}$ predicted by the model. It physically rearranges the patches from their current positions to the target indices, returning the reassembled image $v_{t+1}$.

\section{Details of the Reward Function in RL}
\label{subsec:reward_details}

To effectively guide the policy optimization across diverse tasks, we design a composite reward structure encompassing format constraints $\mathcal{R}_{fmt}$, final accuracy $\mathcal{R}_{acc}$, and intermediate sketch quality $\mathcal{R}_{step}$.

\paragraph{Format Reward Function.}
The format reward $R_{fmt}$ ensures the structural integrity and executability of the model's outputs. Given the hybrid nature of our action space, the model must strictly adhere to the defined protocol: enclosing reasoning in \texttt{<think>} tags and invoking tools via valid \texttt{<tool\_call>} tag with correct argument syntax. We assign a binary penalty as format reward $\mathcal{R}_{\text{fmt}}$:
\begin{equation}
    \mathcal{R}_{\text{fmt}} = 
    \begin{cases} 
    1, & \text{if syntax is valid and executable} \\
    0, & \text{otherwise}
    \end{cases}
\end{equation}
Trajectories violating format constraints are immediately terminated to prevent invalid execution.

\paragraph{Accuracy Reward Function.}
The accuracy reward $R_{\text{acc}}$ evaluates the correctness of the prediction answer $\hat{y}$ against the ground truth answer $y^*$. To accommodate the heterogeneity of tasks in the benchmarks, we employ specific metric definitions corresponding to our implementation:

\begin{itemize}
    \item \textbf{Token-level F1 Score:} For short-answer tasks (e.g., OCR, VQA), we treat the normalized prediction and reference as bags of tokens. We calculate the precision $P$ and recall $R$ based on the intersection of tokens between $\hat{y}$ and $y^*$. The reward is defined as the harmonic mean: $R_{acc-\text{f1}} = \frac{2 \cdot P \cdot R}{P + R}$.
    
    \item \textbf{Exact Match (EM):} For tasks requiring strict output formats (e.g., multiple-choice questions), we employ a binary reward function. $R_{acc-\text{em}} = 1$ if the extracted answer string exactly matches the ground truth after normalization, and $0$ otherwise.
    
    \item \textbf{Array Similarity:} For structured output tasks (e.g., jigsaw puzzles), we quantify the element-wise alignment. Unlike standard accuracy, we penalize length mismatches. Let $\hat{\mathbf{y}}$ and $\mathbf{y}^*$ be the predicted and target integer sequences respectively. The score is calculated as:
    \begin{equation}
        R_{acc-\text{sim}}(\hat{\mathbf{y}}, \mathbf{y}^*) = \frac{\sum_{i=1}^{\min(|\hat{\mathbf{y}}|, |\mathbf{y}^*|)} \mathbb{I}(\hat{y}_i = y^*_i)}{\max(|\hat{\mathbf{y}}|, |\mathbf{y}^*|)},
    \end{equation}
    where $\mathbb{I}(\cdot)$ is the indicator function. This ensures that missing or extra elements reduce the overall score.
    
    \item \textbf{Soft Numerical Reward:} For estimation tasks (e.g., time calculation, instrument reading), we utilize a continuous relaxation of the exact match reward to handle numerical variance. If the absolute error $E = |\hat{y} - y^*|$ is within a small tolerance $\epsilon$, the reward is 1. Otherwise, it follows a decay function based on the normalized error:
    \begin{equation}
        R_{acc-\text{num}}(\hat{y}, y^*) = \frac{1}{1 + \lambda \cdot \frac{E}{\sigma}},
    \end{equation}
    where $\lambda$ is a steepness hyperparameter (set to 10.0), and $\sigma$ is a scaling factor. For time-based tasks, $\sigma$ is fixed at 30 minutes to represent a soft tolerance; for general numerical tasks, $\sigma = \max(|y^*|, 1)$ to scale by the magnitude of the ground truth.

\end{itemize}

\paragraph{Stepwise Reward Function.}
For tasks where the intermediate action parameters have a direct quantifiable correlation with the ground truth, we define precise metric-based rewards. 

For the jigsaw puzzle task, where the action $a_t$ represents a permutation vector of image patch indices, we utilize array similarity to measure the alignment between the current arrangement and the target configuration. This metric quantifies the element-wise accuracy by calculating the proportion of indices that match the ground truth at the same positions.
For the maze navigation task, where $a_t$ corresponds to a sequence of directional moves (e.g., ``U, D, L, R''), we employ the Levenshtein edit distance to quantify how close the generated path is to the optimal solution.
These are formulated as:

\begin{equation}
\begin{split}
  \mathcal{S}_{\text{array}}(a_t, y^*) &= \frac{1}{N} \sum_{i=1}^{N} \mathbb{I}(a_{t,i} = y^*_i), \\
  \mathcal{S}_{\text{edit}}(a_t, y^*) &= 1 - \frac{\text{Lev}(a_t, y^*)}{\max(|a_t|, |y^*|)},
\end{split}
\end{equation}
where $a_{t,i}$ and $y^*_i$ denote the $i$-th element of the current and target permutation vectors respectively, $N$ represents the length of the vector, and $\mathbb{I}(\cdot)$ is the indicator function which equals 1 if the condition holds and 0 otherwise. $\text{Lev}(\cdot, \cdot)$ represents the Levenshtein distance function, normalized to $[0, 1]$ to ensure the reward scale is consistent.

\begin{wraptable}{r}{0.5\textwidth}
\vspace{-10pt}
\centering
\caption{Ablation on the different MLLM evaluators in the RL reward.}
\label{tab:mllm_eval}
\resizebox{\linewidth}{!}{
\begin{tabular}{l|c|cc}
\toprule
Evaluator & TIR-Bench Avg. & Maze & Jigsaw \\
\midrule
\rowcolor{green!10} GPT-4o & 51.8 & 85.0 & 21.2 \\
Qwen3-32B & 50.2 & 85.0 & 21.0 \\
Qwen2.5-7B & 48.0 & 83.2 & 21.2 \\
\bottomrule
\end{tabular}
}
\vspace{-10pt}
\end{wraptable}

\paragraph{Influence of the MLLM evaluator.} We additionally compare the results of employing different MLLMs as the stepwise evaluator, as presented in Table~\ref{tab:mllm_eval}. While our default configuration utilizes GPT-4o, substituting it with medium-scale models still yields competitive performance. Notably, the results under the Qwen3-VL-32B setting are on par with our default metrics, demonstrating that our stepwise reward mechanism is robust to the choice of evaluator when the evaluator itself possesses sufficient reasoning capability. However, when the evaluator's intrinsic capacity is limited (e.g., Qwen2.5-VL-7B), the accuracy of stepwise reward signals degrades, leading to a noticeable drop in overall performance.

\section{Details of Prompts}
We exhibit partial prompts for the VLM evaluation in stepwise reward and thought generation in SFT data.
\begin{promptbox}{Prompts for VLM Evaluation in the Stepwise Reward}
Here is the conversation history:\placeholder{history\_context}
The correct answer (Ground Truth) is: \placeholder{Ans}.

Please examine the provided image carefully \placeholder{img\_url}, and judge whether this historical context and current image provide sufficient, clear, and unambiguous evidence to directly derive the answer? Strictly follow the output format:

 - Do NOT explain or generate any thinking process.

 - Answer immediately with exactly one word: 'Yes' or 'No'.

 Your Answer:

\end{promptbox}

\begin{promptbox}{Prompts for Thought Generation in the SFT Data}

\textbf{Task A: Jigsaw}

You are an expert visual puzzle solver working on restoring a \placeholder{grid\_n}*\placeholder{grid\_n} shuffled image.

 - Current State: \placeholder{state\_curr}
 
 - Next State: \placeholder{state\_next}
 
 - Your Task: Analyze the visual changes to determine if the proposed move improves the image integrity. Please provide a cohesive reasoning paragraph covering the following points naturally:(1) Global Context: Briefly mention the image size and the current tile arrangement \placeholder{state\_curr}. (2)Visual Defects: Identify what looks wrong in the current input (e.g., disconnected lines, fragmented objects, misaligned borders).(3) Restoration Logic: Explain how the proposed rearrangement fixes these specific visual discontinuities. (4)Verification: Conclude which specific region or object is now correctly formed.

Finally, output the target index list clearly as: Target State: \placeholder{state\_next}

\rule{\textwidth}{0.4pt}
    
\textbf{Task B: Maze}

You are an expert in maze path planning. Based on the provided maze information, please thoroughly analyze the current path choices and deduce the optimal move for the next step.

 - Maze Environment Configuration

 This is a \placeholder{grid\_n}*\placeholder{grid\_n} maze map. Coordinates are given in the format (row number, column number).
The red ball is the start point \placeholder{start\_coor}, and the green ball is the endpoint \placeholder{end\_coor}. Black squares represent impassable walls.
The directions `D', `U', `L', and `R' represent moving one step Down, Up, Left, and Right, respectively.

 - Current Path Status

 Your current position (the startpoint) is: \placeholder{cur\_coor}.
The coordinate you need to move to as the correct route (the endpoint of the red line in the second image) is \placeholder{next\_coor}.
Future waypoints are: \placeholder{mid\_coor\_list}.

Please start your reply immediately with a fluent reasoning process. Your deduction must contain the environment description and satisfy the output format constraints. Your response will be a single, continuous paragraph and will not use any subheadings.


\end{promptbox}


\section{Additional Details on Training Data}

\begin{figure*}[h]
  \centering
  \includegraphics[width=\textwidth]{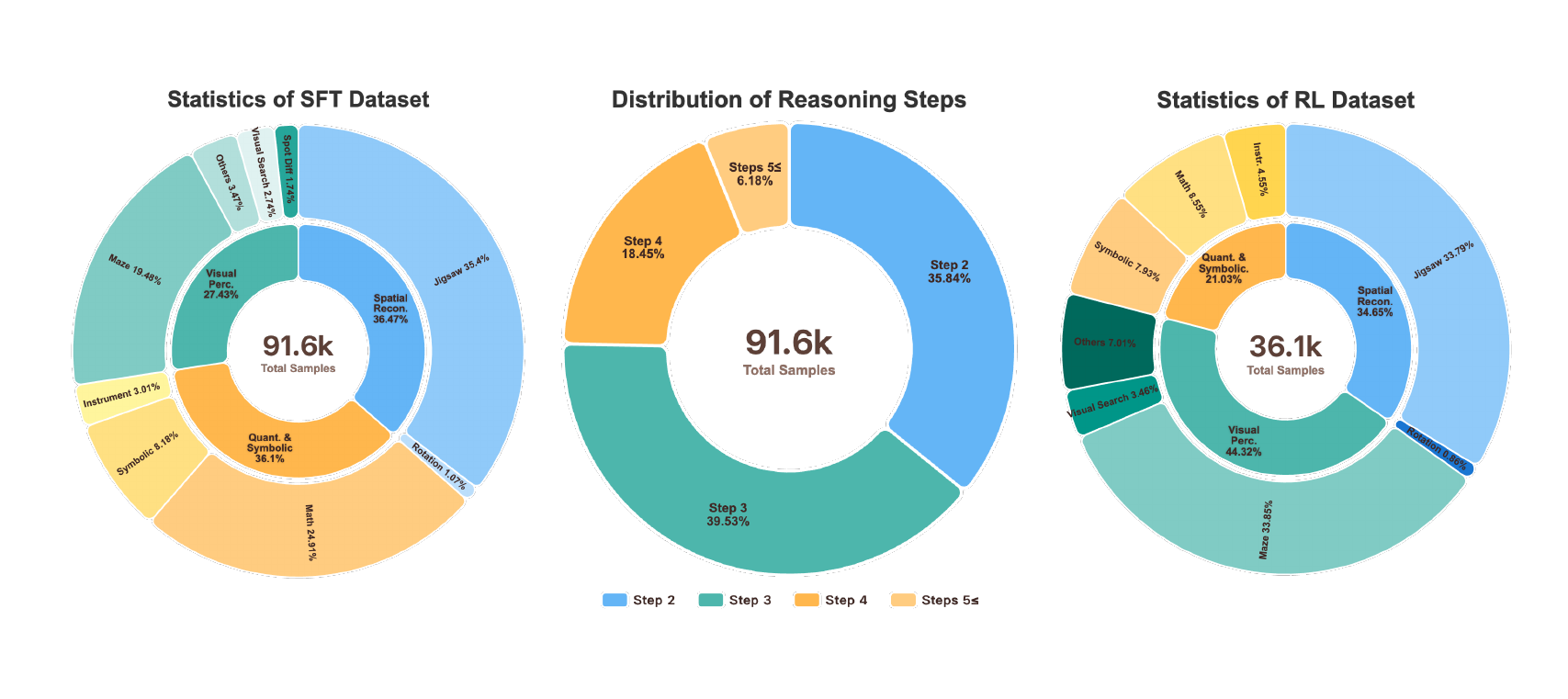}
  \caption{Data statistics for the SFT and RL training stages.}
  \label{fig:data_statistics}
\end{figure*}


Figure~\ref{fig:data_statistics} presents the statistical profile of our training data across three primary cognitive domains: \textbf{Spatial Reconstruction}, \textbf{Visual Perception}, and \textbf{Quantitative Reasoning}. The sunburst charts (left and right) demonstrate a balanced and diverse distribution for both SFT and RL stages, covering a wide array of sub-tasks such as \textit{Jigsaw}, \textit{Maze}, and \textit{Math}. Crucially, the central chart highlights the \textbf{reasoning complexity} of our dataset: the distribution is dominated by multi-step problems, with over 64\% of samples requiring 3 or more steps. This composition ensures the model develops robust, long-horizon reasoning capabilities alongside broad task generalization.

\section{Case Study}
\begin{figure*}[t]
  \centering
  \includegraphics[width=0.95\textwidth]{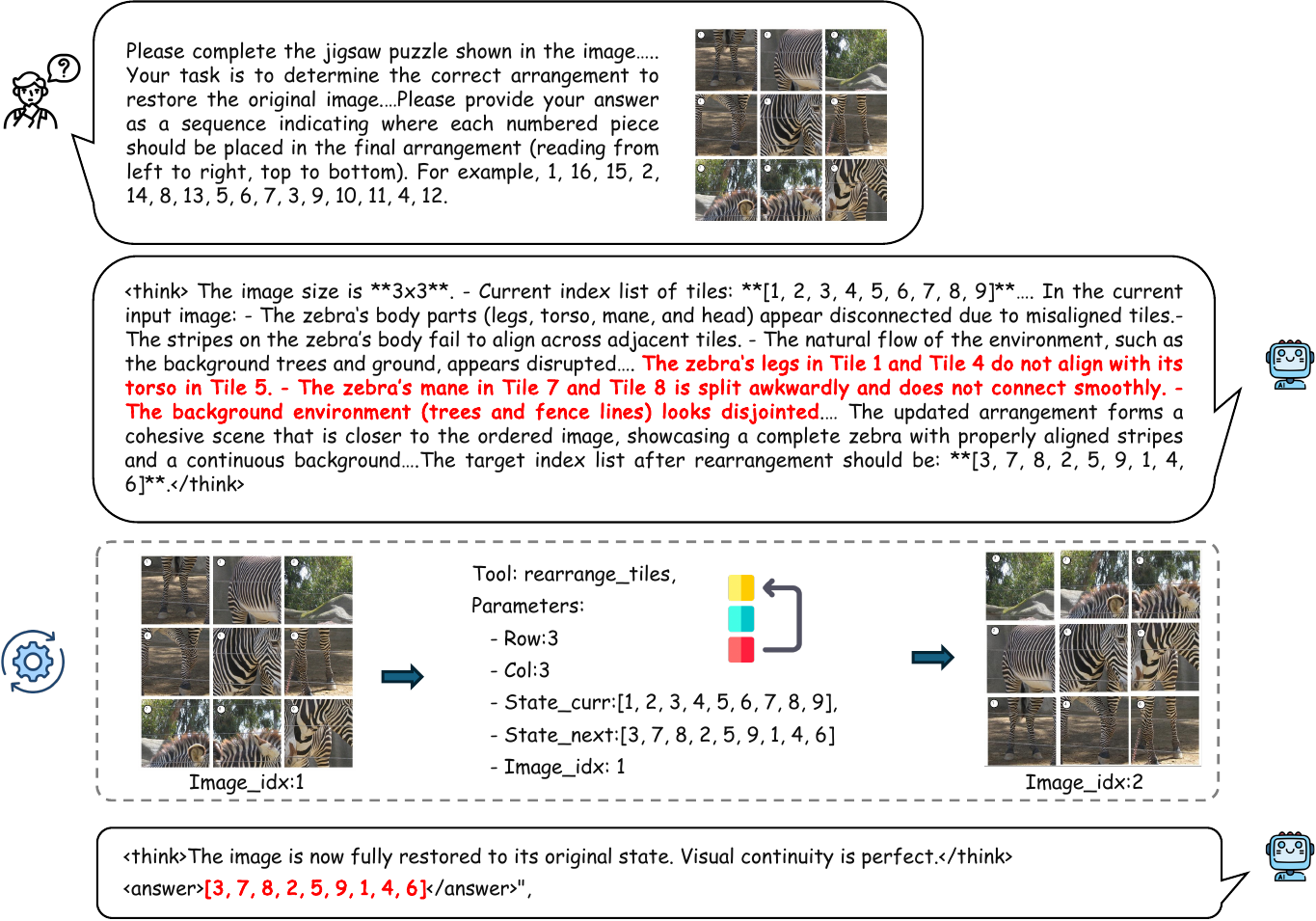}
  \caption{Case study of \ours{} on Jigsaw Game.}
  \label{fig:goodcase_jigsaw_case1}
\end{figure*}

\begin{figure*}[t]
  \centering
  \includegraphics[width=0.9\textwidth]{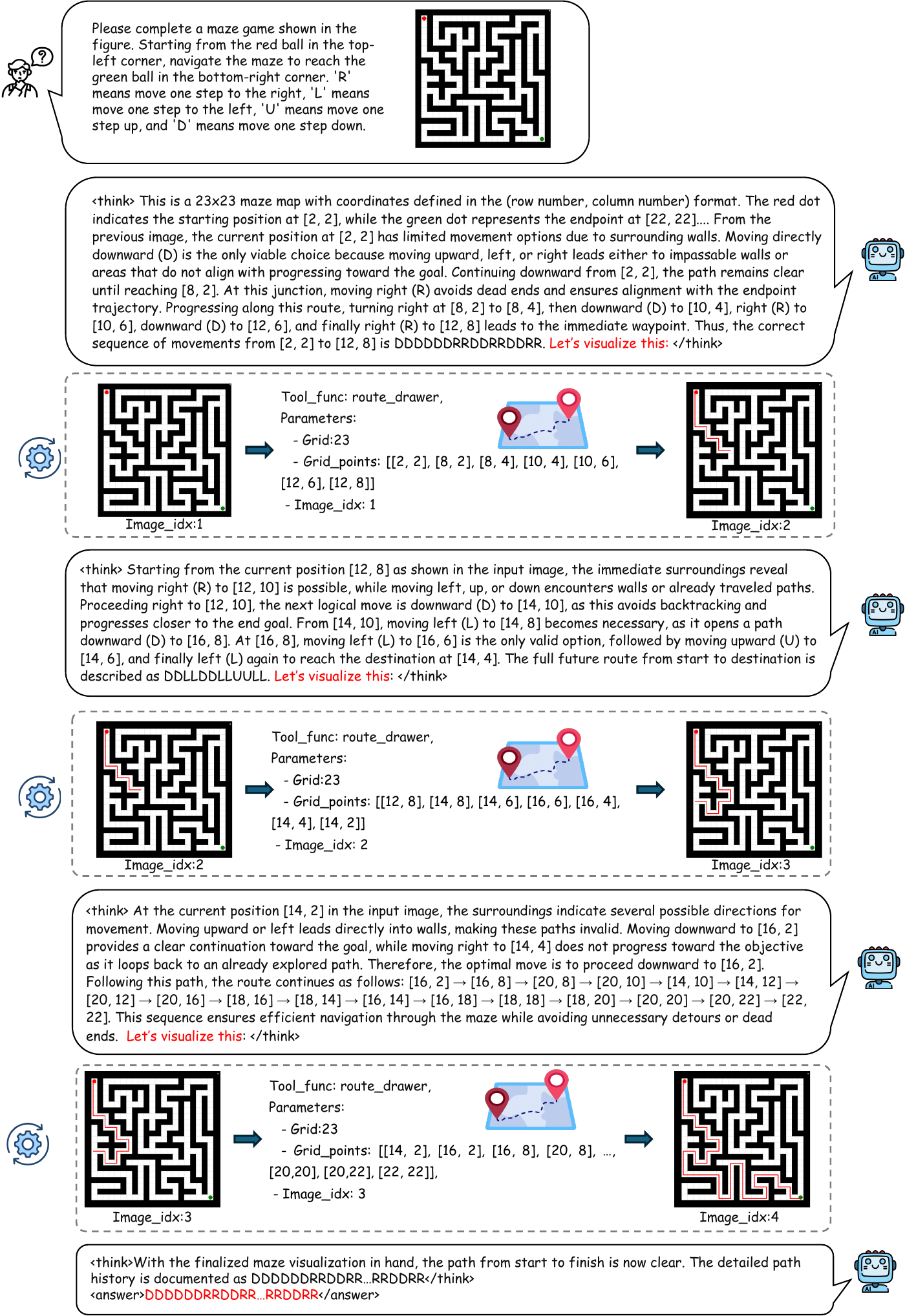}
  \caption{Case study of \ours{} on Maze.}
  \label{fig:goodcase_maze_case1}
\end{figure*}

\begin{figure*}[t]
  \centering
  \includegraphics[width=0.95\textwidth]{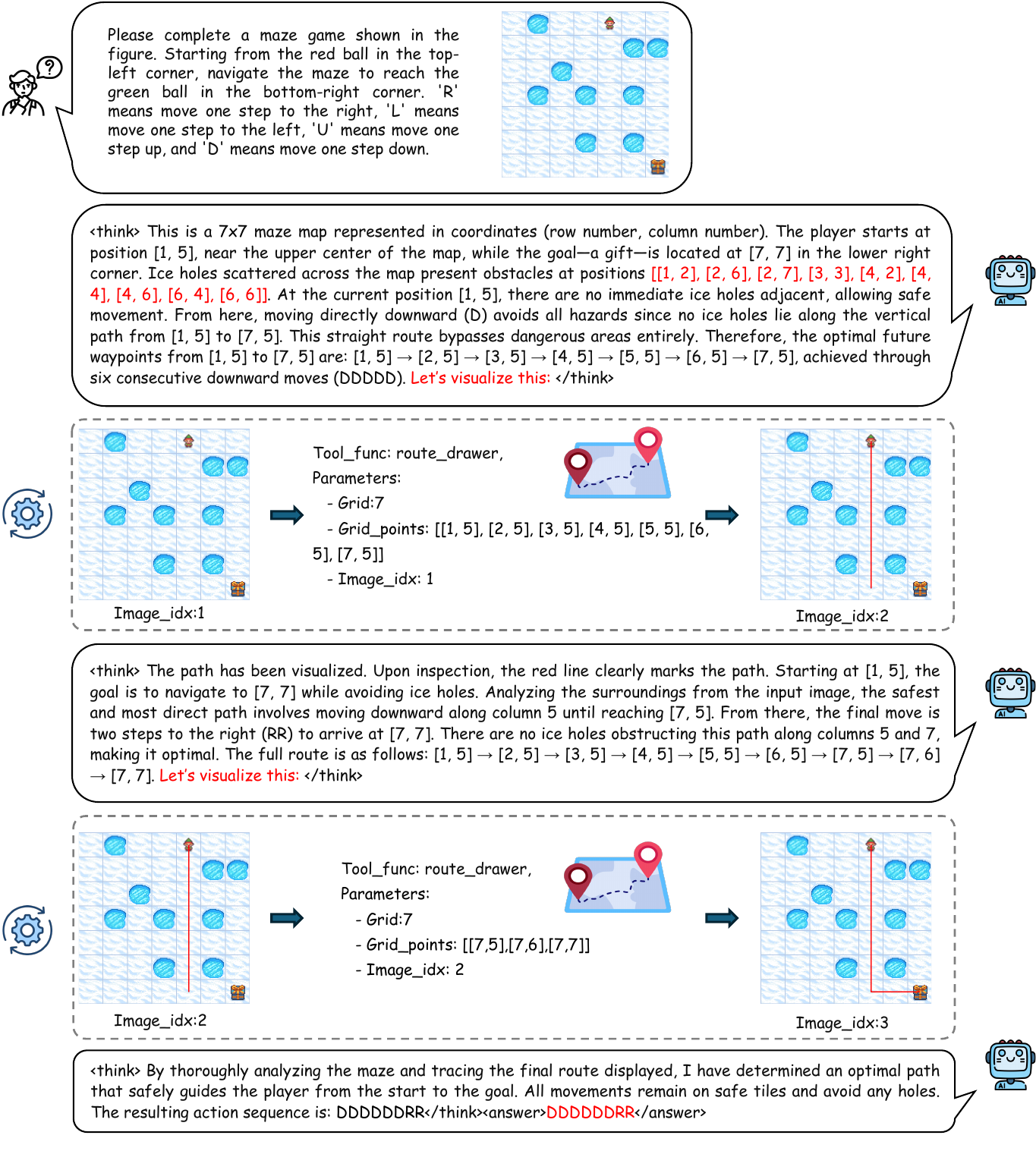}
  \caption{Case study of \ours{} on VSP.}
  \label{fig:goodcase_vsp_case1}
\end{figure*}

\begin{figure*}[t]
  \centering
  \includegraphics[width=0.92\textwidth]{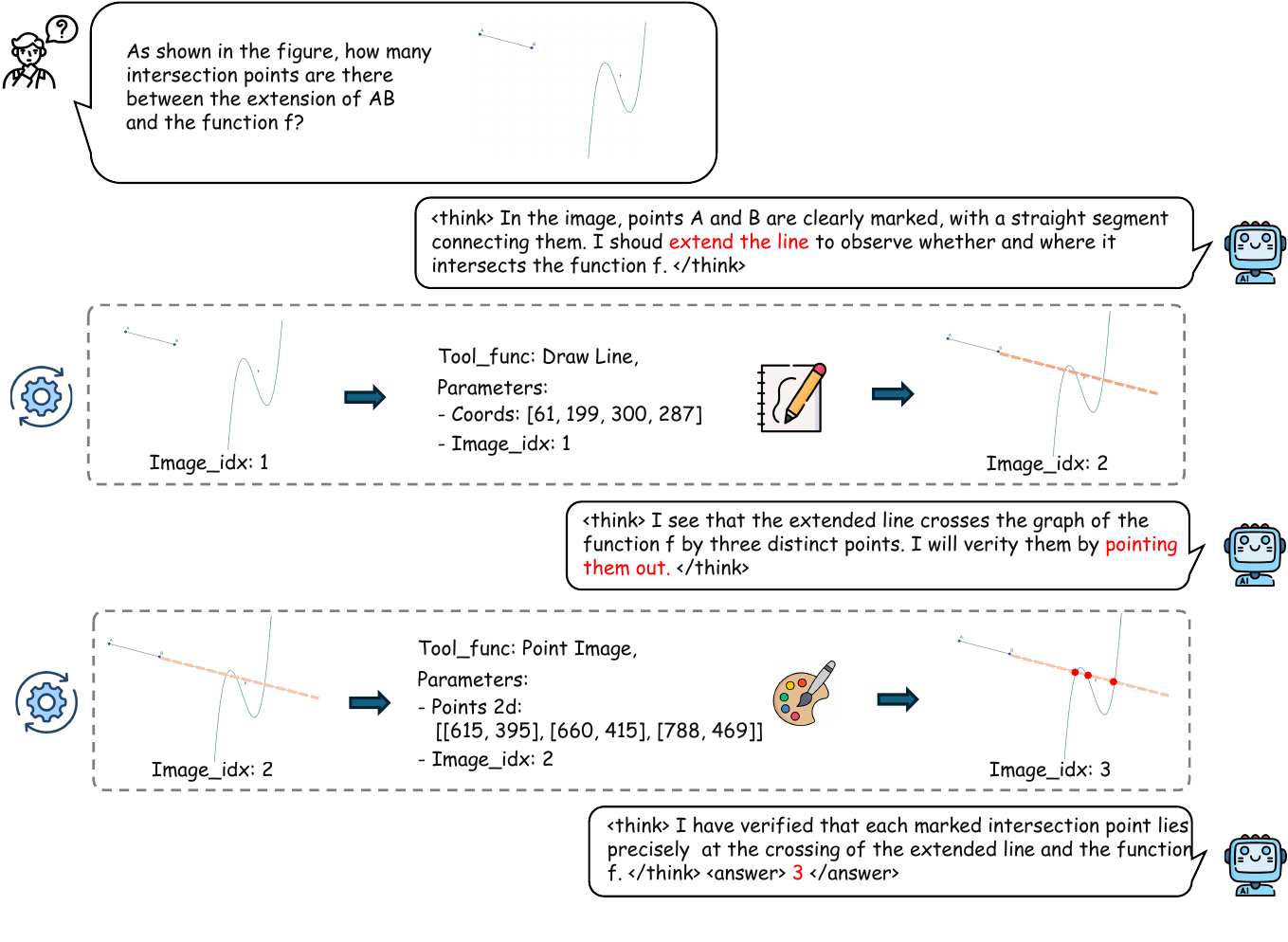}
  \caption{Case study of \ours{} on Math.}
  \label{fig:goodcase_math_case1}
\end{figure*}

\begin{figure*}[t]
  \centering
  \includegraphics[width=0.95\textwidth]{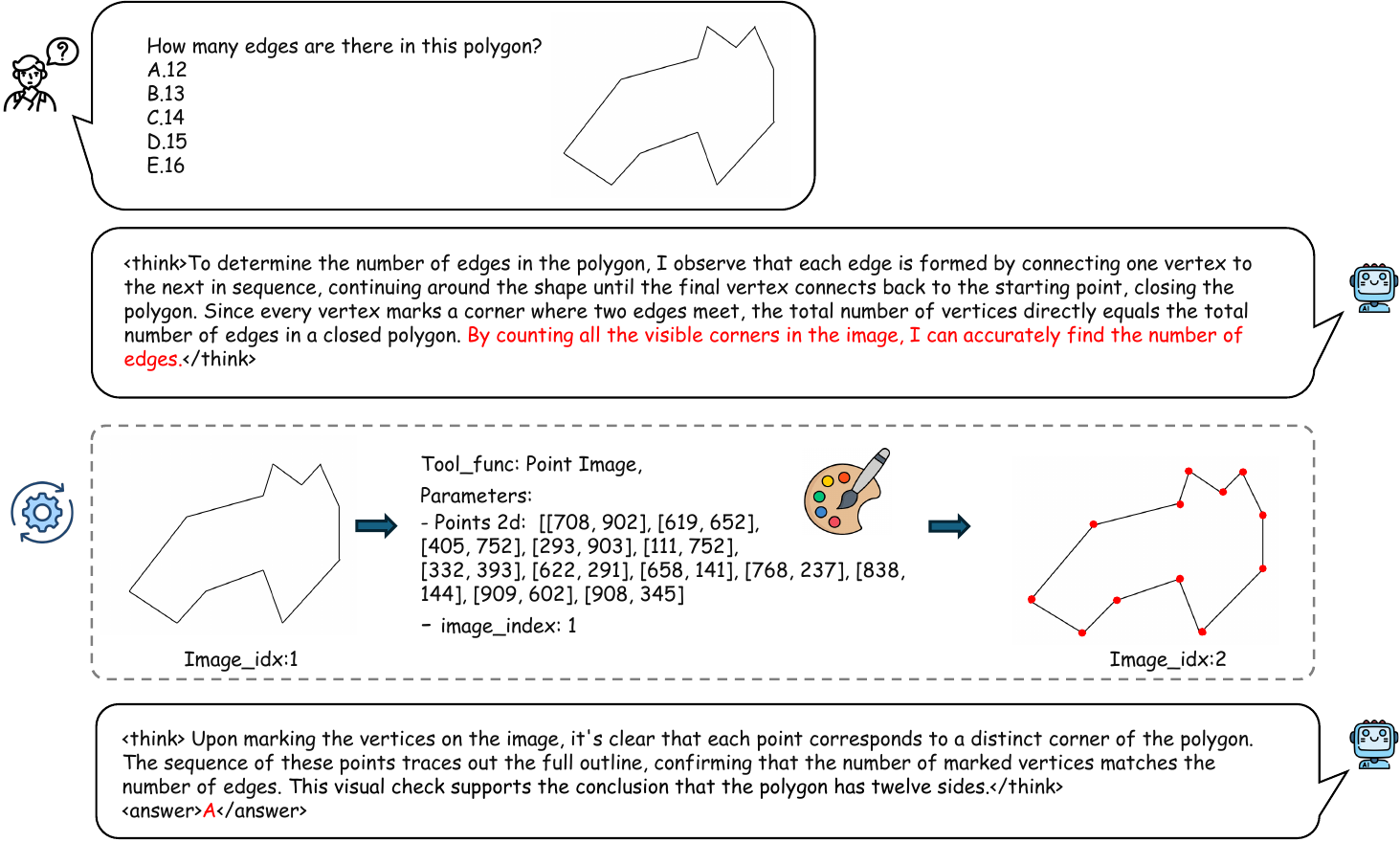}
  \caption{Case study of \ours{} on Symbolic Reasoning.}
  \label{fig:goodcase_symbolic_case1}
\end{figure*}

\begin{figure*}[t]
  \centering
  \includegraphics[width=0.95\textwidth]{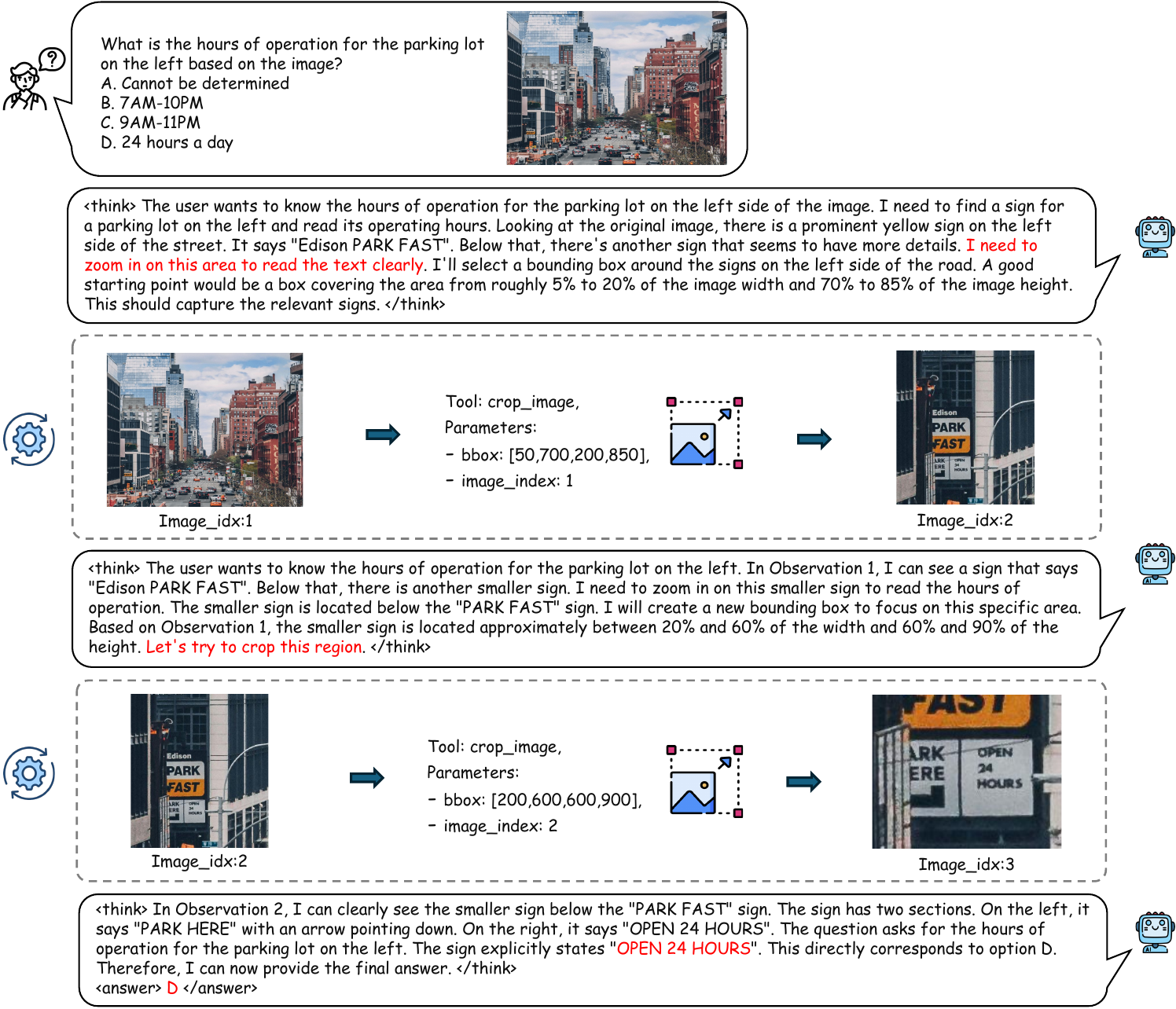}
  \caption{Case study of \ours{} on Visual Search.}
  \label{fig:goodcase_visual_search_case1}
\end{figure*}

\begin{figure*}[t]
  \centering
  \includegraphics[width=0.95\textwidth]{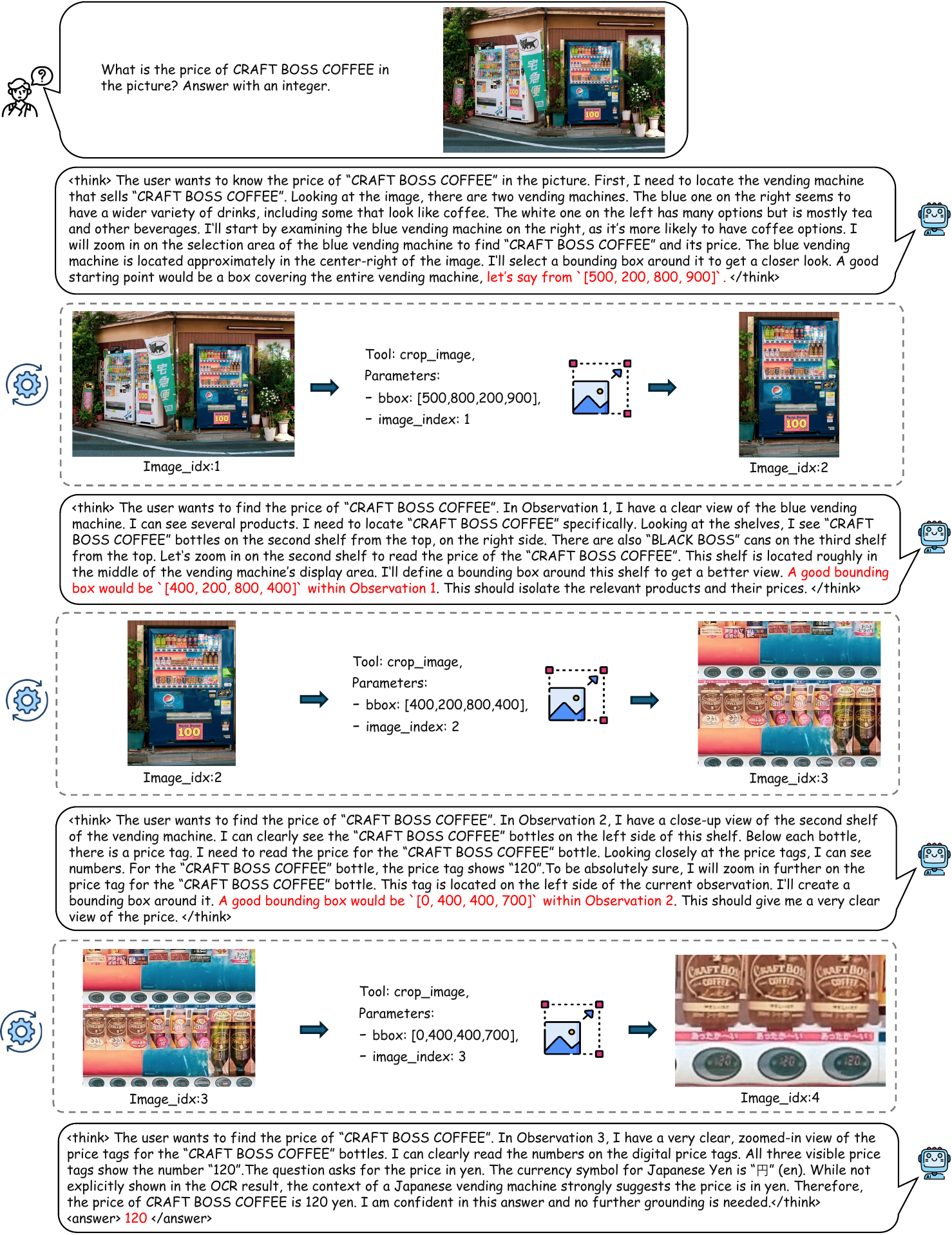}
  \caption{Case study of \ours{} on Visual Search.}
  \label{fig:goodcase_visual_search_case2}
\end{figure*}

\begin{figure*}[t]
  \centering
  \includegraphics[width=0.95\textwidth]{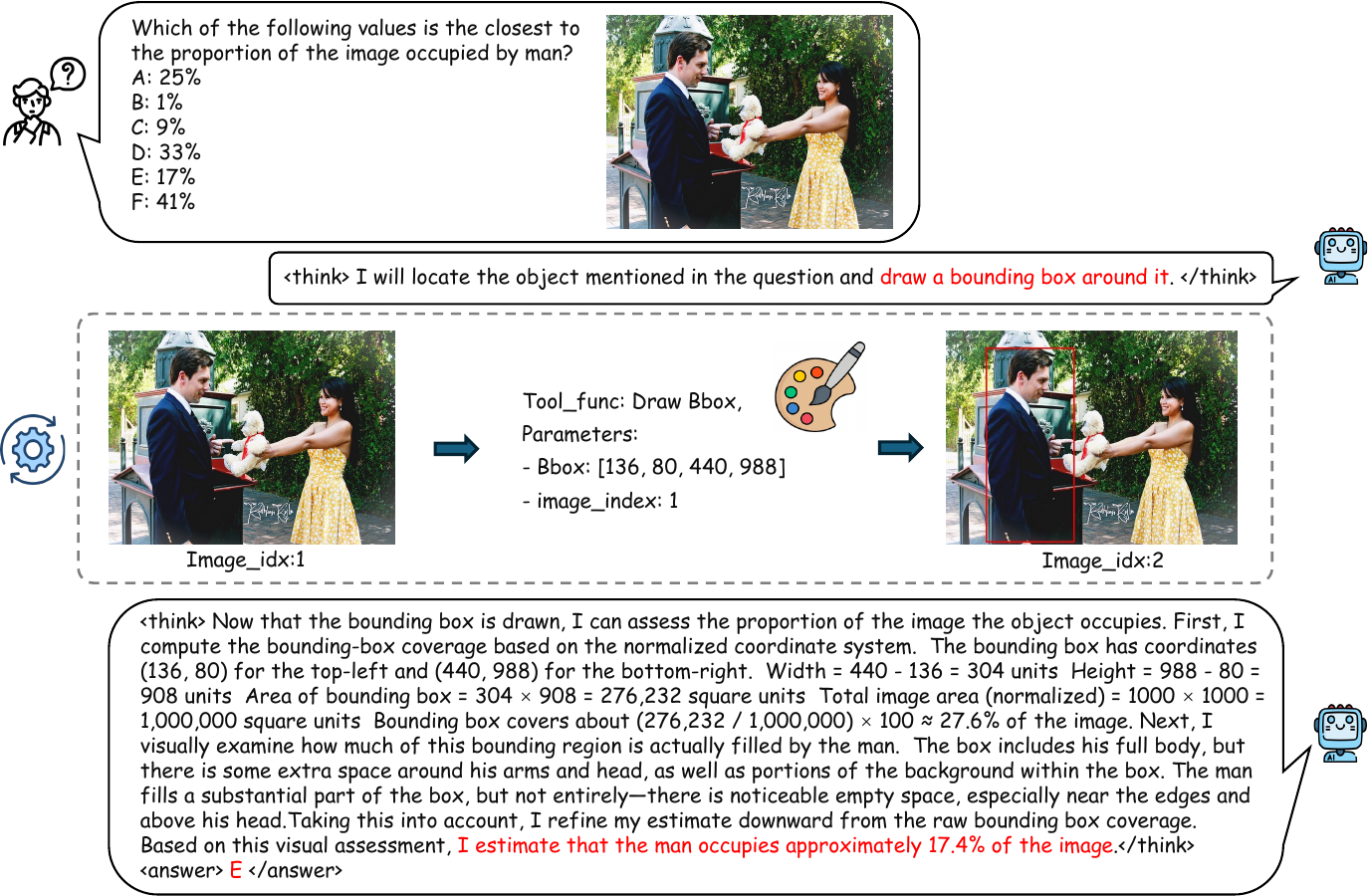}
  \caption{Case study of \ours{} on Proportion VQA.}
  \label{fig:goodcase_proportion_vqa_case2}
\end{figure*}

\begin{figure*}[t]
  \centering
  \includegraphics[width=0.95\textwidth]{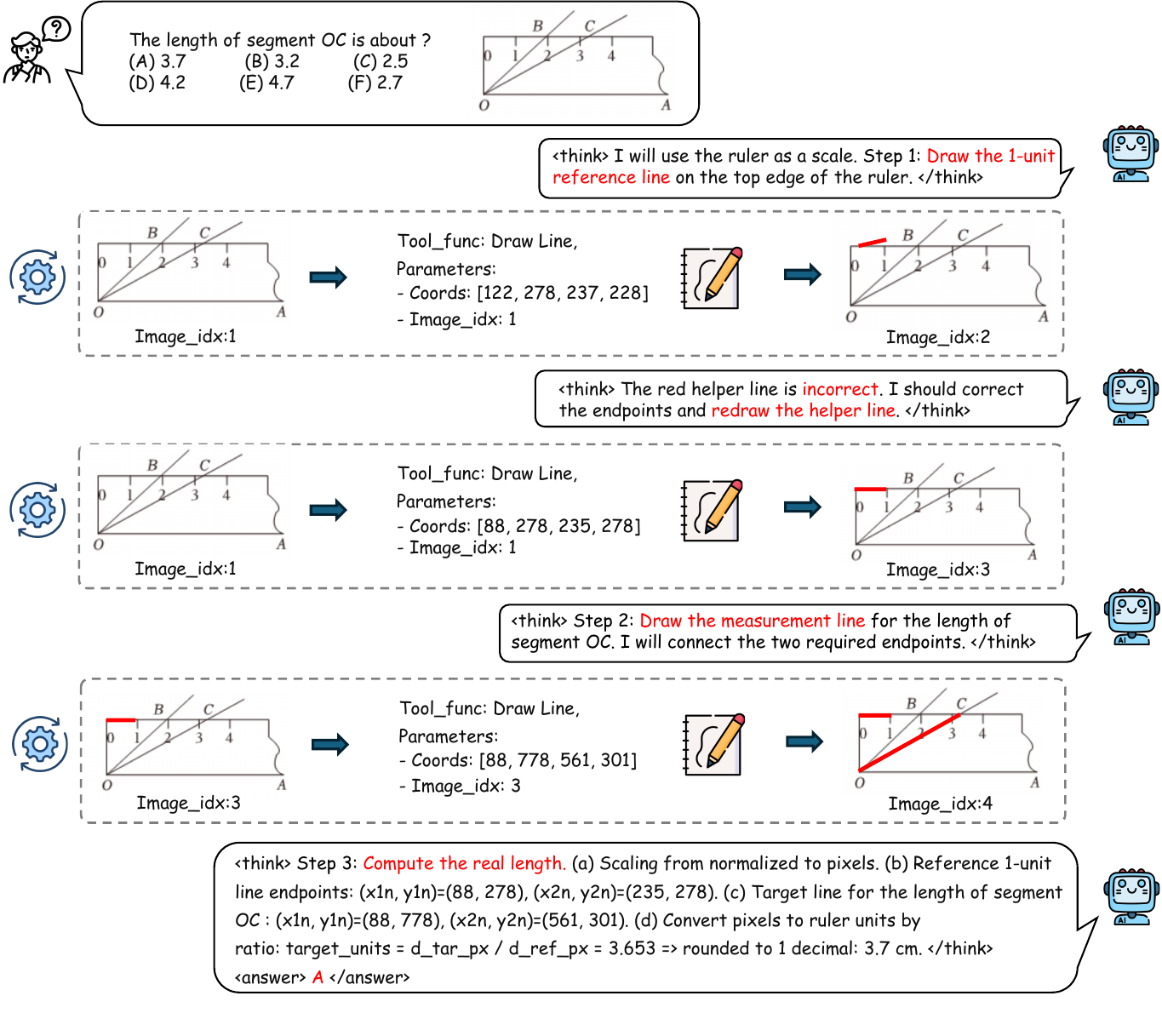}
  \caption{Case study of \ours{} on Math with reflection.}
  \label{fig:goodcase_math_case1_reflection}
\end{figure*}

\begin{figure*}[t]
  \centering
  \includegraphics[width=0.95\textwidth]{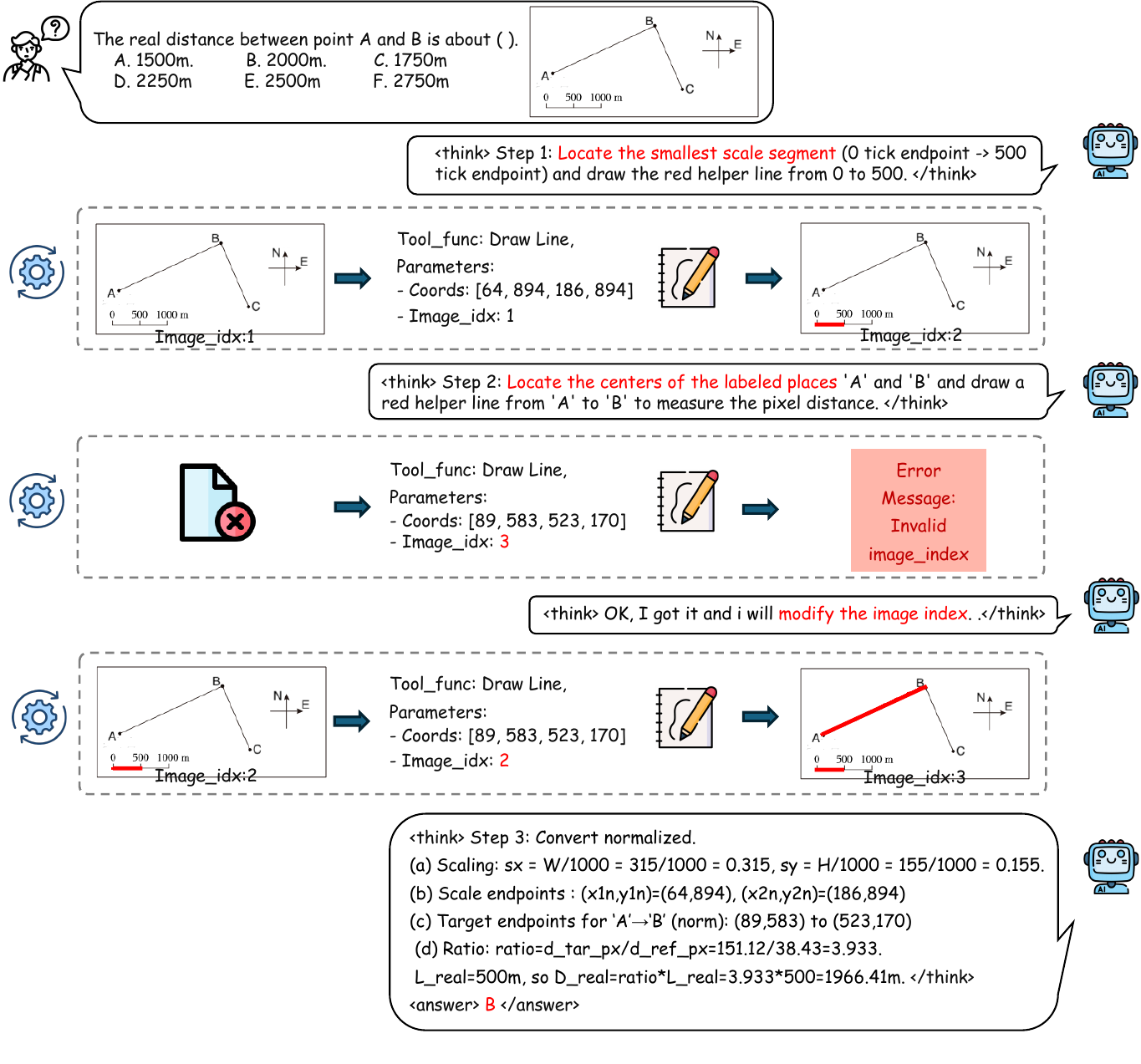}
  \caption{Case study of \ours{} on Math  with reflection.}
  \label{fig:goodcase_math_case2_reflection}
\end{figure*}

We showcase a diverse set of examples demonstrating the robust performance of \ours{}, including the maze task (Figure~\ref{fig:goodcase_maze_case1}), the VSP task (Figure~\ref{fig:goodcase_vsp_case1}), the math task (Figure~\ref{fig:goodcase_math_case1}), the symbolic reasoning task (Figure~\ref{fig:goodcase_symbolic_case1}), the jigsaw game (Figure~\ref{fig:goodcase_jigsaw_case1}), the visual search(Figures~\ref{fig:goodcase_visual_search_case1} and \ref{fig:goodcase_visual_search_case2}), and proportion VQA (Figure~\ref{fig:goodcase_proportion_vqa_case2}).
As illustrated in Figures~\ref{fig:goodcase_math_case1_reflection} and \ref{fig:goodcase_math_case2_reflection}, \ours{} learns to reflect on previous mistakes such as an incorrect action, and accordingly revises its subsequent decision when solving the math problem. 
This error-aware adjustment demonstrates a human-like reflection behavior, enabling the model to diagnose failure modes in its reasoning process and recover by generating a more accurate next action.

\begin{figure*}[t]
  \centering
  \includegraphics[width=0.95\textwidth]{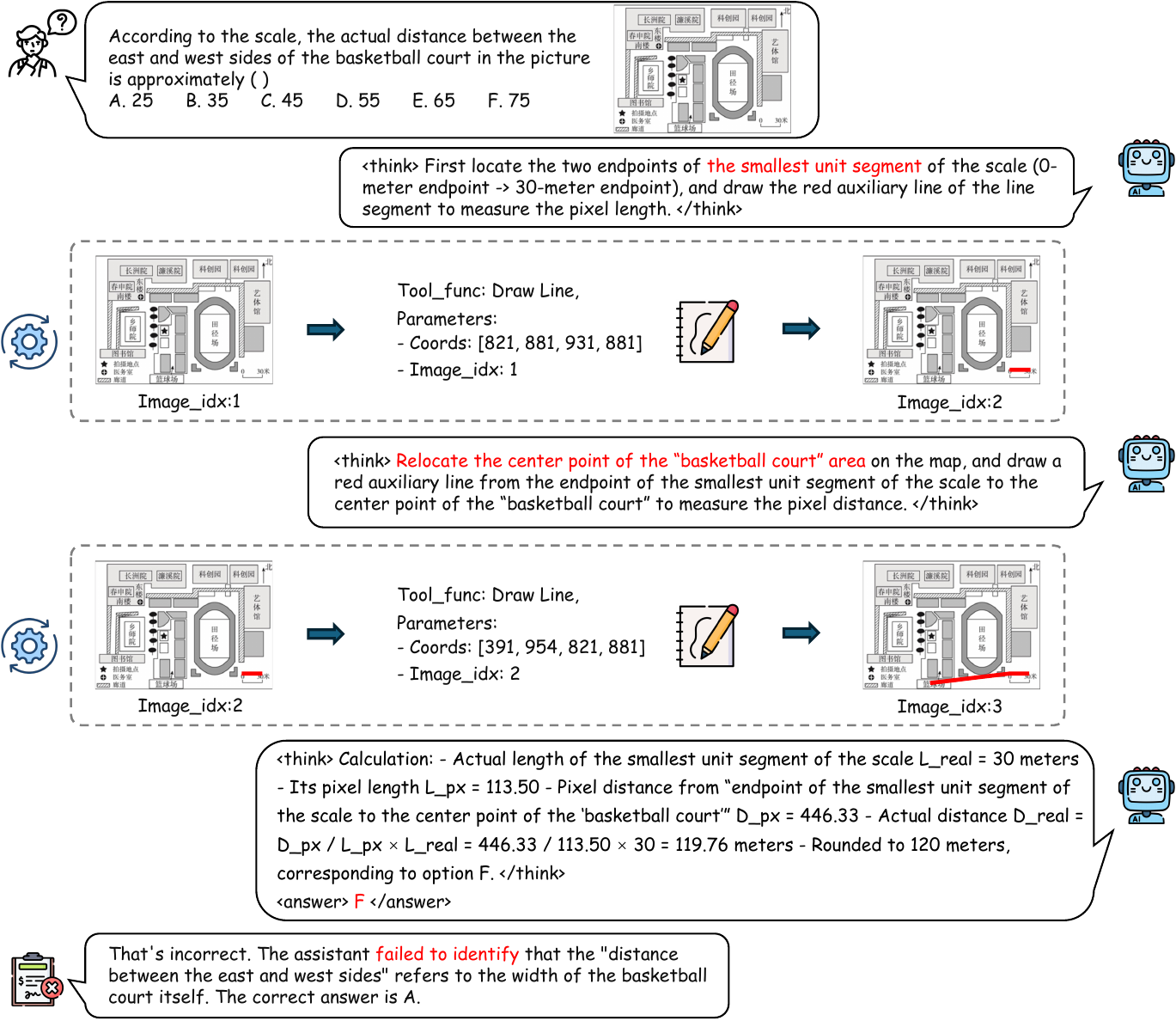}
  \caption{Failure case study of \ours{} on Math.}
  \label{fig:badcase_math_case2}
\end{figure*}

\begin{figure*}[t]
  \centering
  \includegraphics[width=0.95\textwidth]{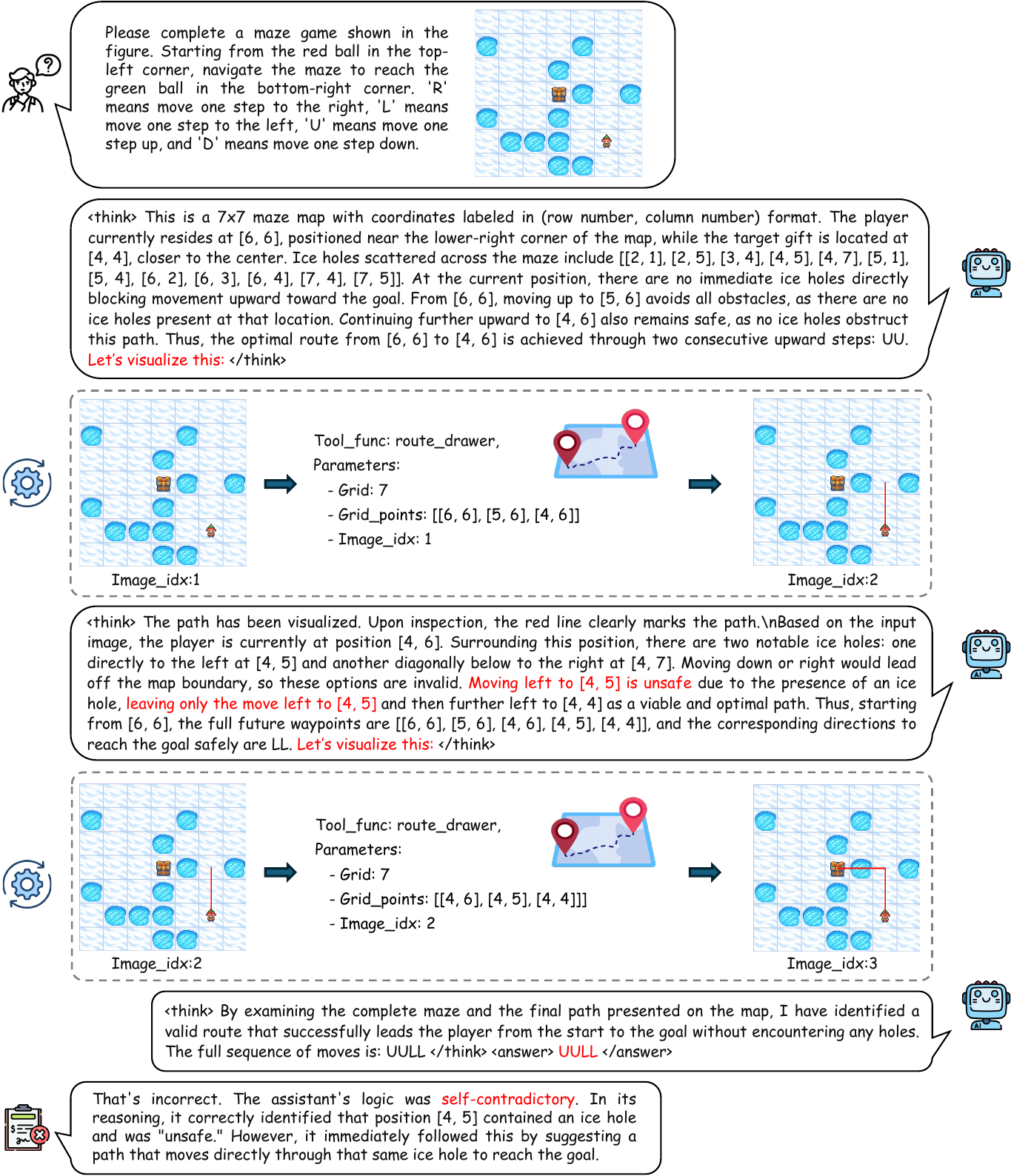}
  \caption{Failure case study of \ours{} on VSP.}
  \label{fig:badcase_maze_case1}
\end{figure*}




\end{document}